\documentclass[12pt]{article}
\usepackage{epsfig, epsf, graphicx, subfigure}
\usepackage{pstricks, pst-node, psfrag}
\usepackage{amssymb,amsmath,amsthm,bm}
\usepackage{verbatim,enumerate}
\usepackage{rotating, lscape,natbib}

\usepackage{setspace}
\AtBeginDocument{%
	\addtolength\abovedisplayskip{-0.2\baselineskip}%
	\addtolength\belowdisplayskip{-0.2\baselineskip}%
}

\def\diag{\hbox{diag}}
\def\wh{\widehat}

\def\diag{\hbox{diag}}

\def\boxit#1{\vbox{\hrule\hbox{\vrule\kern6pt
			\vbox{\kern6pt#1\kern6pt}\kern6pt\vrule}\hrule}}

\def\bse{\begin{eqnarray*}}
	\def\ese{\end{eqnarray*}}
\def\be{\begin{eqnarray}}
\def\ee{\end{eqnarray}}
\def\bq{\begin{equation}}
\def\eq{\end{equation}}
\def\bse{\begin{eqnarray*}}
	\def\ese{\end{eqnarray*}}

\def\tr{\hbox{tr}}

\def\wh{\widehat}

\newcommand{\bI}{\mathbf{I}}
\newcommand{\bX}{\mathbf{X}}
\newcommand{\bK}{\mathbf{K}}

\newcommand{\bx}{\mathbf{x}}

\newcommand{\bPhi}{\mathbf{\Phi}}
\newcommand{\bPsi}{\mathbf{\Psi}}

\newcommand{\bA}{\mathbf{A}}
\newcommand{\bW}{\mathbf{W}}
\newcommand{\bD}{\mathbf{D}}
\newcommand{\bB}{\mathbf{B}}

\newcommand{\EE}{\mathbb{E}_{\varepsilon}}

\newcommand{\Vare}{{\rm Var}_{\varepsilon}}
\newcommand{\Cove}{{\rm Cov}_{\varepsilon}}

\newcommand{\Peps}{\mathbb{P}_{\varepsilon}}
\newcommand{\Pb}{\mathbb{P}}
\newcommand{\Pepsx}{\mathbb{P}_{\varepsilon,X}}
\newcommand{\Pyx}{\mathbb{P}_{Y,X}}
\newcommand{\Px}{\mathbb{P}_{X}}
\newcommand{\ePx}{\mathbb{P}}
\newcommand{\VE}{\hbox{Var}_{\varepsilon}}
\newcommand{\GCV}{\hbox{GCV}}
\newcommand{\dGCV}{\hbox{dGCV}}

\newcommand{\beps}{\boldsymbol{\varepsilon}}

\newcommand{\bbeta}{\boldsymbol{\beta}}
\newcommand{\btheta}{\boldsymbol{\theta}}
\def\Xv{\mathbf X}
\def\Yv{\mathbf Y}
\def\xv{\mathbf x}
\def\yv{\mathbf y}
\def\fv{\mathbf f}
\def\Fv{\boldsymbol F}

\newtheorem{theorem}{Theorem}

\newtheorem{lemma}{Lemma}
\newtheorem{remark}{Remark}

\begin{document}
	\newcommand{\xggrev}[1]{{\color{magenta}{#1}}}
	
	\thispagestyle{empty} \baselineskip=28pt \vskip 5mm
	\begin{center} {\huge{\bf Distributed Generalized Cross-Validation for  Divide-and-Conquer Kernel Ridge Regression and its Asymptotic Optimality}}
	\end{center}

	\baselineskip=11pt \vskip 10mm
	
	\begin{center}\large
Ganggang Xu\footnote[1]{
	\baselineskip=10pt Department of Management Science,
  University of Miami, Coral Gables, 33146, USA. \\
	E-mail: gangxu@bus.miami.edu}, 
Zuofeng Shang\footnote[2]{
	\baselineskip=10pt Department of Mathematical Sciences,
	Indiana University-Purdue University Indianapolis, Indianapolis, IN 46202, USA. \\
	E-mail: shangzf@iu.edu},
	and Guang Cheng\footnote[3]{
	\baselineskip=10pt  Department of Statistics,
	Purdue University, West Lafayette, IN 47907-2066
	USA.\\
	E-mail: chengg@stat.purdue.edu}
	\footnote[4]{This paper is an extended version of the work published in the conference paper \cite{XSC18}. We extend our previous work in methodology (Sections 2.2, 3.5), theory (Section 4.2) and simulation studies (Sections 5.1.2-5.2.2).}
	\end{center}
	
	\baselineskip=17pt \vskip 10mm \centerline{\today} \vskip 10mm

	\begin{center}
		{\large{\bf Abstract}}
    \end{center}
    Tuning parameter selection is of critical importance for kernel ridge regression. To this date, data driven tuning method for divide-and-conquer kernel ridge regression (d-KRR) has been lacking in the literature, which limits the applicability of d-KRR for large data sets.  In this paper, by modifying the Generalized Cross-validation \citep[GCV,][]{Wahba90} score, we propose a distributed Generalized Cross-Validation (dGCV) as a data-driven tool for selecting the tuning parameters in d-KRR. Not only the proposed dGCV is computationally scalable for massive data sets, it is also shown, under mild conditions, to be asymptotically optimal in the sense that minimizing the dGCV score is equivalent to minimizing the true global conditional empirical loss of the averaged function estimator, extending the existing optimality results of GCV to the divide-and-conquer framework. 

\section{Introduction}\label{sec:intro}
Massive data made available in various research areas have imposed new challenges for data scientists. With a large to massive sample size, many sophisticated statistical tools are no longer applicable simply due to formidable computational costs and/or memory requirements. Even when the computation is possible on more advanced machines, it is still appealing to develop accurate statistical procedures at much lower computational costs. { The divide-and-conquer strategy has become a popular tool for regression models. With carefully designed algorithms, such a strategy has proven to be effective in Linear models \citep{CX14,LCL16}, Partially linear models \citep{ZCL16} and Nonparametric regression models~\citep{ZDW15,LGZ17,SC17,GSW17}. In this paper, we shall focus on the divide-and-conquer kernel ridge regression (d-KRR) where the selection of the penalty parameter is of vital importance but still remains unsettled.}

Suppose we have independent and identically distributed samples $\{(x_i,y_i)\in\mathcal{X}\times \mathbb{R}\}_{i=1,\dots N}$ from a joint probability measure $\Pyx$. The goal is to study the association between the covariate vector $\bm x_i$ and the response $y_i$ through the following model
\be
\label{model}
y_i=f_0(\bm x_i)+\varepsilon_i,\quad i=1,\dots,N,
\ee
where $f_0(\cdot):\mathcal{X}\to \mathbb{R}$ is the function of interest and $\varepsilon_i$ is a random error term with mean zero and a common variance $\sigma^2$. One popular method to estimate $f_0(\cdot)$ is the {\em Kernel Ridge Regression} \citep{ShCr04} which essentially aims at finding a projection of $f_0(\cdot)$ into a reproducing kernel Hilbert space (RKHS), denoted as $\mathcal H$, equipped with a norm $\|\cdot\|_{\mathcal H}$. Specifically, the KRR estimator is then defined as
\be
\label{pls1}
\widehat f=\arg\min_{f\in\mathcal{H}}\left\{\frac{1}{N}\sum_{i=1}^N(y_i-f(\bm x_i))^2+\lambda\|f\|_{\mathcal{H}}^2\right\},
\ee
where $\lambda\geq 0$ controls trade-off between goodness-of-fit and smoothness of $f$. 

It is well known that computing $\widehat f$ requires $O(N^3)$ floating operations and $O(N^2)$ memory; see \eqref{pls2} for more details. When $N$ is large, such requirements can be prohibitive. To overcome this, \citet{ZDW15} proposed the following ``divide-and-conquer'' algorithm: (i) Randomly divide the entire sample $\{(\bm x_1, y_1),\dots,(\bm x_N, y_N)\}$ to $m$ disjoint ``smaller'' subsets, denoted by $S_1,\dots,S_m$; (ii) For each subset $S_k$, find $
\widehat f_k=\arg\min_{f\in\mathcal{H}}\left\{\frac{1}{n_k}\sum_{i\in S_k}(y_i-f(\bm x_i))^2+\lambda\|f\|_{\mathcal{H}}^2\right\}$,
where $n_k$ is the size of $S_k$; (iii) The final nonparametric estimator is given by
\be
\label{ave}\bar{f}(x)=\frac{1}{m}\sum_{k=1}^m\widehat f_k(\bm x).
\ee
Such a ``divide-and-conquer'' strategy reduces computing time from $O(N^3)$ to $O(N^3/m^2)$ and memory usage from $O(N^2)$ to $O(N^2/m^2)$. Both savings may be substantial as $m$ grows. Furthermore, \citet{ZDW15} shows that as long as $m$ does not grow too fast, the averaged estimator $\bar{f}$ achieves the same minimax optimal estimation rate as the oracle estimate $\widehat f$, i.e., (\ref{pls1}), that utilizes all data points at once. In this sense, the divide-and-conquer algorithm is quite appealing as it achieves an ideal balance between the computational cost and the statistical efficiency.  

However, the aforementioned statistical efficiency depends critically on a careful choice of tuning parameter $\lambda$ in all sub-samples. The optimal choice of tuning parameter $\lambda$ has been well studied for KRR when the entire data set can be fitted at once. Examples include Mallow's CP \citep{Ma73}, Generalized cross-validation \citep[GCV,][]{CW79} and Generalized approximated cross-validation \citep{XW96}. However, if we naively apply these traditional tuning methods in each sub-sample to pick an optimal $\lambda_k$ in the above step (ii), the averaged function estimator $\bar{f}$ subsequently obtained using~(\ref{ave}) will be sub-optimal. As pointed out by existing literature \citep[e.g.][]{ZDW15, BM17, CLZ17}, the optimal tuning parameter should be chosen in accordance with the order of {\em the entire sample size}, i.e., $N$, such that we intentionally allow the resulting sub-estimator $\widehat f_k$ to over-fit the sub-sample $S_k$ for each $k=1,\dots,m$. { Based on the order of the optimal choice of $\lambda$, \citet{ZDW15} proposed a heuristic data-driven approach to empirically choose an optimal $\lambda$. However, the theoretical properties of this approach remain unclear.}  In this paper, we define a new data-driven criterion named ``distributed generalized cross-validation" (dGCV) to choose tuning parameters for KRR under the divide-and-conquer framework. The computational cost of the proposed criterion remains the same as $O(N^3/m^2)$. More importantly, we show that the proposed method enjoys similar theoretical optimality as the well-known GCV criterion \citep{CW79} in the sense that the resulting divide-and-conquer estimate minimizes the true empirical loss function asymptotically.

The rest of paper are organized as follows.	Section~\ref{sec:krr} introduces background on kernel ridge regression. Section~\ref{sec:dgcv} presents the main result of this paper on the dGCV, while Section~\ref{sec:thry} gives statistical guarantee for this new tuning procedure. Our method and theory are backed up by extensive simulation studies in Sections~\ref{sec:sim}, and are applied to the Million Song Dataset in Section~\ref{sec:rea}, demonstrating significant advantages over \cite{ZDW15}. All technical proofs are postponed to the Appendix.

\section{Kernel Ridge Regression Estimation}\label{sec:krr}

In this section, we briefly review kernel ridge regression \citep{ShCr04}. The reproducing kernel Hilbert space, denoted as $\mathcal{H}$, is a Hilbert space induced by a symmetric nonnegative definite kernel function $K(\cdot,\cdot):\mathcal{X}\times \mathcal{X}\to \mathbb{R}$ and an inner product $\langle\cdot,\cdot\rangle_{\mathcal{H}}$ satisfying
\[
\langle g(\cdot),K(\bm x,\cdot)\rangle_{\mathcal{H}}=g(\bm x)\text{ for any }g\in\mathcal{H}.
\]
The kernel function $K(\cdot,\cdot)$ is called the reproducing kernel of the Hilbert space $\mathcal{H}$ equipped with the norm $\|g\|_{\mathcal{H}}=\sqrt{\langle g(\cdot),g(\cdot)\rangle_{\mathcal{H}}}$.  Using the Mercer's theorem, under some regularity conditions, the kernel function $K(\cdot,\cdot)$ possesses the expansion $K(\bm x,\bm z)=\sum_{j=1}^{\infty}\mu_j\psi_j(\bm x)\psi_j(\bm z)$,
where $\mu_1\geq\mu_2\geq\dots$ is a sequence of decreasing eigenvalues and $\{\psi_1(\cdot),\psi_2(\cdot),\dots\}$ is a family of orthonormal basis functions of $L^2(\Px)$. The smoothness of $g \in \mathcal{H}$ is characterized by the decaying rate of the eigenvalues $\{\mu_j\}_{j=1}^\infty$. There are three types of estimation considered in this paper, including smoothing spline \citep{Wahba90} as a special case. 

{\bf Finite rank:} There exists some integer $r$ such that $\mu_j = 0$ for $j>r$. For example, with vectors $\bm x,\bm z$, the polynomial kernel $K(\bm x,\bm z) = (1+\bm x^T\bm z)^r$ has a finite rank $r+1$, and induces a space of polynomial functions with degree at most $r$. This corresponds to the parametric ridge regression. 

{\bf Exponentially decaying:} There exist some $\alpha, r>0$ such that $\mu_j \asymp \exp(-\alpha j^r)$. Exponentially decaying kernels include the multivariate Gaussian kernel $K(\bm x,\bm z) = \exp(-\|\bm x-\bm z\|_2^2/\phi^2)$, where $\phi>0$ is the scale parameter and $\|\cdot\|_2$ is the Euclidean norm.

{\bf Polynomially decaying:} There exists some $r>0$ such that $\mu_j \asymp j^{-2r}$. The polynomially decaying class includes many smoothing spline kernels of the Sobolev space \citep{Wahba90}. For example, kernel function $K(x,z)=1+\min(x,z)$ induces the Sobolev space of Lipschitz functions with smoothness $\nu=1$ and has polynomially decaying eigenvalues.

\subsection{The Representer Theorem}
With observed data, using the representer theorem \citep{Wahba90}, it can be shown that the solution to the minimization problem (\ref{pls1}) takes the following form
\be
\label{sol1}
\widehat f(\bm x)=\sum_{i=1}^N\beta_iK(\bm x_i,\bm x),
\ee
where $\beta_1,\dots,\beta_N\in \mathbb{R}$. Furthermore, based on the observed sample, the parameter vector $\bbeta=(\beta_1,\dots\beta_N)^T$ can be estimated by minimizing the following criterion
\be
\label{pls2}
\frac{1}{N}(\Yv-\bbeta^T\bK)^T(\Yv-\bbeta^T\bK)+\lambda\bbeta^T\bK\bbeta,
\ee
where $\Yv=(y_1,\dots,y_N)^T$ and $\bK=[K(\bm x_i,\bm x_j)]_{i,j=1,\dots,N}$. The solution to \eqref{pls2} takes the form of $\wh\bbeta=(\bK+N\lambda\bI_N)^{-1}\Yv$, which requires $O(N^3)$ operations.

We next apply the above idea to sub-estimation. Denote $(\yv_1,\xv_1),\dots,(\yv_m,\xv_m)$ as a random partition of the entire data with $\yv_k=(y_{k,1},\dots,y_{k,n_k})^T$ and $\xv_k=(\bm x_{k,1},\dots,\bm x_{k,n_k})^T$. Define vectors $\fv_k=(f_0(\bm x_{k,1}),\dots,f_0(\bm x_{k,n_k}))^T$ and $\beps_k=\yv_k-\fv_k$. Define the sub-kernel matrices $\bK_{kl}=\left[K(\bm x_{i},\bm x_{j})\right]_{i\in S_k,j\in S_l}$ for $l,k=1,\dots,m$. It is straightforward to show that the minimizer of~(\ref{pls2}) with $\bK$ replaced by $\bK_{kk}$ is of the form $\wh\bbeta_k=(\bK_{kk}+n_k\lambda\bI_k)^{-1}\yv_k$, and the individual function estimator $\widehat f_k(x)$ can be written as
\be
\label{sol-1}
\widehat f_k(\bm x)=\sum_{i\in S_k}\widehat\beta_{k,i}K(\bm x_{i},\bm x),
\ee
where $\widehat\beta_{k,i}$ is the entry of $\wh\bbeta_k$ corresponding to $x_{k,i}$, $k=1,\dots,m$.  
\subsection{Kernel Ridge Regression for Multivariate Functions}
\label{sec:add}
In principle, any multivariate function~$f_0(\bm x)$ in~\eqref{model}, i.e., $\bm x\in\mathbb{R}^p$, can be well approximated if a sufficiently good reproducing kernel $K(\cdot,\cdot)$ can be identified. However, for a large $p$, the excessive risk of the KRR estimator may grow exponentially fast as the dimension $p$ increases \citep{gyorfi2006distribution}, which is often referred to as the ``curse of dimensionality".  One common strategy is to impose some special structures on the reproducing kernel. For example, the polynomial kernel $K(\bm x,\bm z) = (1+\bm x^T\bm z)^r$ assumes that $K(\cdot,\cdot)$ depends only on the inner product of $\bm x$ and $\bm z$ and the multivariate Gaussian kernel $K(\bm x,\bm z) = \exp(-\|\bm x-\bm z\|_2^2/\phi^2)$ assumes that $K(\cdot,\cdot)$ is determined by the Euclidean distance between vectors $\bm x$ and $\bm z$. More sophisticated applications of Gaussian kernels may also allow the scale parameter $\phi$ to vary for different dimensions. Another popular approach to circumvent the ``curse of dimensionality" is to use additive approximation \citep{hastie2017generalized, kandasamy2016additive} to multivariate functions. Let $\bm x=(x_1,x_2,\cdot,x_p)^T$, and define the first-order additive approximation of $f(\bm x)$ as 
\be
\label{add}
f^*(\bm x)=f_1^*(x_1)+\cdots+f_p^*(x_p),
\ee
where each $f_j^*(\cdot)$ is a univariate function residing in a reproducing kernel Hilbert space $\mathcal{H}_k$ with a reproducing kernel $K_j(\cdot,\cdot)$, $j=1,\cdots,p.$ The corresponding additive kernel can be defined as
$K(\bm x,\bm z)=\sum_{j=1}^{p}K_j(x_j,z_j)$,
and the associated reproducing kernel Hilbert space is $\mathcal{H}=\mathcal{H}_1\bigoplus\mathcal{H}_2\bigoplus\cdots\bigoplus\mathcal{H}_p$. For some applications where the first order approximation~\eqref{add} is not adequate, higher order additive approximations to the multivariate function $f(\bm x)$ can be used to achieve better estimation accuracies at similar computational costs, see \cite{kandasamy2016additive} for more detailed discussions.

\section{Tuning Parameter Selection}\label{sec:dgcv}
\subsection{Sub-GCV Score: Local Optimality}\label{sec:4_2}

In this section, we define the GCV score for each sub-estimation, named as sub-GCV score, and discuss its theoretical property. 
Define the empirical loss function for $\widehat f_k$ as follows
\be
\label{lossk}
L_k(\lambda|\xv_k)=\frac{1}{n_k}\sum_{i\in S_k}w_{i}\left\{\widehat f_k(x_{i})-f_0(x_{i})\right\}^2,
\ee
where $w_{i}\geq 0$ is some weight assigned to each observation $(y_{i},x_{i})$ and satisfies $\sum_{i\in S_k}w_{i}=n_k$. The introduction of weights in (\ref{lossk}) helps reducing computational cost; see Section~\ref{sec:cc}.  The tuning parameter $\lambda$ is referred to as ``locally optimal" if it only minimizes local empirical loss $L_k(\lambda|\xv_k)$. When only focused on a single sub-data set, such a ``locally-optimal" choice of tuning parameter $\lambda$ has been well studied in \citep{CW79, Li86, Gu02, Wood04, GM05, XH12}, among which the Generalized Cross-Validation \citep{CW79} remains to be one of the most popular approaches. 

Using the function estimator $\widehat f_k(x)$, the predicted values for the vector $\yv_k$ can be written as
$\wh\yv_k=\bA_{kk}(\lambda)\yv_k$, where $\bA_{kk}(\lambda)=\bK_{kk}(\bK_{kk}+n_k\lambda\bI_k)^{-1}$. 
Here the matrix $\bA_{kk}(\lambda)$ is often known as the hat matrix. Using the above notations, the sub-GCV score is defined as
\be
\label{gcv1}
\GCV_k(\lambda)=\frac{n_k^{-1}(\wh\yv_k-\yv_k)^T\bW_k(\wh\yv_k-\yv_k)}{\{1+n_k^{-1}\tr\{\bA_{kk}(\lambda)\bW_k\}\}^2},
\ee
where $\bW_k=\diag\{w_{i},i\in S_k\}$, $k=1,\dots,m.$
It is well known that $\GCV_k(\lambda)$ enjoys appealing asymptotic properties. For example, under mild conditions, \citet{Gu02} showed that, as $n_k\to\infty$,
\[
\GCV_k(\lambda)-L_k(\lambda|\xv_k)-\frac{1}{n_k}\beps_k^T\bW_k\beps_k=o_{\Peps}\{L_k(\lambda|\xv_k)\}, 
\]
$k=1,\dots,m.$  This property essentially asserts that, minimizing $\GCV_k(\lambda)$ with respect to $\lambda$ is asymptotically equivalently to minimizing the local ``golden criterion" $L_k(\lambda|\xv_k)$.
\subsection{Local-Optimality v.s. Global-Optimality}\label{sec:2}
\label{locvsglob}

In this section, we explain why the use of $\GCV_k(\lambda)$ in each subsample does not lead to an optimal averaged estimate $\bar f$.
We first derive conditional risks for both $\widehat f_k$ and $\bar f$. For the former, some basic algebra yields that the conditional risk $R_k(\lambda|\xv_k)=\EE\left\{L_k(\lambda|\xv_k)\right\}$ is of the form
\be\label{riskk}
\begin{split}
	R_k(\lambda|\xv_k)&=\frac{1}{n_k}\sum_{i\in S_k}w_i\VE\left\{\widehat f_k(x_{i})\right\}+\frac{1}{n_k}\sum_{i\in S_k}w_i\left\{\EE\widehat f_k(x_{i})-f_0(x_{i})\right\}^2,
\end{split}
\ee
where the expectation is taken with respect to the probability measure $\Peps$. As for the latter, we first define the empirical loss function of  $\bar{f}$ as
\be
\label{loss}
\bar{L}(\lambda|\Xv)=\frac{1}{N}\sum_{i=1}^Nw_i\{\bar{f}(x_i)-f_0(x_i)\}^2,
\ee
where $\Xv=(\xv_1,\dots,\xv_m)$ denotes the collection of all covariates and $w_i\geq 0$ are the associated weights with observation $i$ such that $\sum_{i=1}^{N}w_i=N$. Similarly, the corresponding conditional risk $	\bar{R}(\lambda|\Xv)=\EE\{\bar{L}(\lambda|\Xv)\}$ has the following form 
\be\label{risk}
\begin{split}
	\bar{R}(\lambda|\Xv)&=\frac{1}{N}\sum_{i=1}^Nw_i\left[\frac{1}{m}\sum_{k=1}^m\left\{\EE\widehat{f}_k(x_i)-f_0(x_i)\right\}\right]^2+\frac{1}{m^2N}\sum_{k=1}^{m}\sum_{i=1}^Nw_i\VE\left\{\widehat f_k(x_i)\right\}.
\end{split}
\ee

The form of (\ref{riskk}) illustrates that, roughly speaking, a ``locally optimal" choice of $\lambda$ (that minimizes (\ref{lossk})) tries to strike a good balance of variance and bias for each sub-estimate $\widehat f_k$. On the contrary, a ``globally optimal" $\lambda$, which is defined to minimize (\ref{loss}), puts much less emphasis on the variance of $\widehat f_k$ (by a factor of $1/m$) than on the bias of $\widehat f_k$; see (\ref{risk}). Consequently, to obtain a ``globally optimal" $\bar f$, one needs to intentionally choose a ``smaller" $\lambda$ such that each individual function estimator $\widehat{f}_k$ overfits data set $S_k$, which leads to reduced bias $\EE\widehat{f}_k(x_i)-f_0(x_i)$ and inflated variance $\VE\left\{\widehat f_k(x_i)\right\}$. Then by taking $\bar{f}=\frac{1}{m}\sum_{j=1}^m\widehat{f}_j$, the variance of $\bar{f}$ can be effectively reduced by a factor of $1/m$ while keeping its bias at the same level as those of individual $\widehat{f}_j$'s. The above risk analysis confirms the heuristics in \citet{ZDW15}.

\subsection{Distributed Generalized Cross-Validation}

The discussions in Section \ref{sec:2} motivate the main result of this paper: distributed GCV score, denoted by dGCV. This data-driven tool in selecting $\lambda$ is computationally efficient for massive data as analyzed in Section \ref{sec:cc}.

Using the solution~(\ref{sol-1}), it is straightforward to show that the predicted values of all data points $\yv_l$ in the subset $S_l$ using $\widehat{f}_k$ take the form $\wh\yv_{kl}=\bA_{kl}\yv_k$, where $\bA_{kl}(\lambda)=\bK_{kl}^T(\bK_{kk}+n_k\lambda\bI_k)^{-1}$. Define the pooled vector of responses $\Yv=(\yv_1^T,\dots,\yv_m^T)^T$. Then the predicted value of $\Yv$ using the averaged estimator $\bar{f}$ is of the form $$\wh\Yv=\left(\frac{1}{m}\sum_{k=1}^{m}\wh\yv_{k1}^T,\dots,\frac{1}{m}\sum_{k=1}^{m}\wh\yv_{km}^T\right)^T=\bar{\bA}_m(\lambda)\Yv,$$ where the averaged hat matrix $\bar{\bA}_m(\lambda)$ is defined as follows 
\be
\label{Am}
\begin{split}
\bar{\bA}_m(\lambda)&=\frac{1}{m}\left(
\begin{array}{cccc}
\bA_{11}(\lambda) & \bA_{12}(\lambda)  & \cdots &\bA_{1m}(\lambda) \\
\bA_{21}(\lambda) & \bA_{22}(\lambda)  & \cdots &\bA_{2m}(\lambda) \\
\vdots & \vdots  & \ddots & \vdots \\
\bA_{m1}(\lambda) & \bA_{m2}(\lambda)  & \cdots &\bA_{mm}(\lambda) \\
\end{array}
\right)
\end{split}.
\ee
Furthermore, the global conditional risk function~(\ref{risk}) can be conveniently re-written as
\be
\label{risk2}
\begin{split}
	\bar{R}(\lambda|\Xv)&= \frac{1}{N}\Fv^T\{\bI-\bar{\bA}_m(\lambda)\}^T\bW\{\bI-\bar{\bA}_m(\lambda)\}\Fv+\frac{\sigma^2}{N}\tr\left\{\bar{\bA}_m^T(\lambda)\bW\bar{\bA}_m(\lambda)\right\},
\end{split}
\ee
where vector of true values $\Fv=(\fv_1^T,\dots,\fv_m^T)^T$ and $\bW=\diag\{w_1,\dots,w_N\}$. Obviously the risk function above cannot be used to select $\lambda$ in practice since the vector $\Fv$ is unknown. Following \citet{Gu02}, we can define an unbiased estimator of $\bar{R}(\lambda|\Xv)+\sigma^2$ as follows
\be
\label{U}
\begin{split}
	\bar{U}(\lambda|\Xv)&=\frac{1}{N}\Yv^T\{\bI-\bar{\bA}_m(\lambda)\}^T\bW\{\bI-\bar{\bA}_m(\lambda)\}\Yv+\frac{2\sigma^2}{N}\tr\left\{\bar{\bA}_m(\lambda)\bW\right\}.
\end{split}
\ee
It is straightforward to show that $\EE\{\bar{U}(\lambda|\Xv)\}=\bar{R}(\lambda|\Xv)+\sigma^2$. The above $\bar{U}(\lambda|\Xv)$ can be viewed as an extension of the Mallow's CP \citep{Ma73} to the divide-and-conquer scenario. 

Similar to \citet{Gu02, XH12}, the Lemma~\ref{lem1} in Section \ref{sec:thry} states that under some mild conditions, minimizing $\bar{U}(\lambda|\Xv)$ and $\bar{L}(\lambda|\Xv)$ with respect to $\lambda$ is asymptotically equivalent. In this sense, the $\lambda$ chosen by minimizing $\bar{U}(\lambda|\Xv)$ is therefore ``globally optimal." However, a major drawback of $\bar{U}(\lambda|\Xv)$ is that it utilizes the knowledge of $\sigma^2$, which in practice often needs to be estimated. To overcome this, we propose the following modification of the GCV score
\be
\label{gcv-m}
\dGCV(\lambda|\Xv)=\frac{\frac{1}{N}\sum_{i=1}^Nw_i\left\{y_i-\bar f(x_i)\right\}^2}{\left[1-\frac{1}{Nm}\sum_{k=1}^m\tr\{\bA_{kk}(\lambda)\bW_k\}\right]^2},
\ee
where $\bW_k=\diag\{w_i,i\in S_k\}$. Intuitively, consider $\tilde{\sigma}^2=N^{-1}\sum_{i=1}^Nw_i\left\{y_i-\bar f(x_i)\right\}^2$ as an estimator of $\sigma^2$ and use the fact that $(1-x)^{-2}\approx 1+2x$ as $x\to 0$, the $\bar{U}(\lambda|\Xv)$ defined in (\ref{U}) essentially can be viewed as the first order Taylor expansion of the $\dGCV(\lambda|\Xv)$. However, in the definition of $\dGCV(\lambda|\Xv)$, it does not require any information of $\sigma^2$. Note that dGCV incorporates  information across all sub-samples, which explains its superior empirical performance. In fact, Theorem~\ref{thm1} in Section \ref{sec:thry} shows that under some conditions, minimizing $\dGCV(\lambda|\Xv)$ and the ``golden criterion" $\bar{L}(\lambda|\Xv)$ with respect to $\lambda$ are also asymptotically equivalent. 

\subsection{Computational Complexity of dGCV}\label{sec:cc}
The computation of $\dGCV(\lambda|\Xv)$ in (\ref{gcv-m}) for a given $\lambda$ consists of two parts: the first part involves computing the trace of individual hat matrices, $\tr\{\bA_{kk}(\lambda)\bW_k\}$, $k=1,\dots,m$, which requires $O(N^3/m^2)$ floating operations and a memory usage of $O(N^2/m^2)$; the second part is to evaluate the predicted value of $\bar{f}(x_i)$ for which $w_i\neq 0$, which costs $O(NN_w)$ floating operations and a memory usage of $O(N)$, where $N_w$ denotes the number of nonzero $w_i$'s.  Hence, the total computation cost of $\dGCV(\lambda|\Xv)$ is of the order $O(N^3/m^2+NN_w)$. In  cases when $m/\sqrt{N}=O(1)$, one can simply use $w_1=\dots=w_N=1$, which results in the computational cost of the order $O(N^3/m^2)$ for one evaluation of $\dGCV(\lambda|\Xv)$. This is the same as that of the divide-and-conquer algorithm proposed in \citet{ZDW15}. 

In some applications where $m$ is much larger than $\sqrt{N}$, the computational cost of $\dGCV(\lambda|\Xv)$ becomes $O(NN_w)$. In this case, we may want to only choose $m^*$ out of $m$ sub-data sets for saving computational costs. To achieve that, we need to choose weights $w_i$'s properly. For example, we can set $w_i=N/(\sum_{k=1}^{m^*}n_k)$ if $i\in\cup_{k=1}^{m^*}S_k$ and $w_i=0$ otherwise. Under this setting, the $\dGCV(\lambda|\Xv)$ in (\ref{gcv-m}) becomes
\be
\label{gcv-m1}
\dGCV^*(\lambda|\Xv)=\frac{\frac{1}{N_{m^*}}\sum_{i\in\cup_{k=1}^{m^*}S_k}\left\{y_i-\bar f(x_i)\right\}^2}{\left[1-\frac{1}{mN_{m^*}}\sum_{k=1}^{m^*}\tr\{\bA_{kk}(\lambda)\}\right]^2},
\ee
where $N_{m^*}=n_1+\dots+n_{m^*}.$ Using (\ref{gcv-m1}) instead of (\ref{gcv-m}), we only need to evaluate $\bar{f}(x_i)$ for $x_i$'s in $m^*$ subsets and the computation time is reduced to $O(N^2m^*/m+N^3/m^2)$. We applied (\ref{gcv-m1}) to the Million Song Data set considered in Section~6, which yields good results in both prediction and computation time.

Optimization of $\dGCV(\lambda|\Xv)$ or $\dGCV^*(\lambda|\Xv)$ can be carried out using a simple one-dimensional grid search. Since the first and second derivatives of $\dGCV(\lambda|\Xv)$ or $\dGCV^*(\lambda|\Xv)$ can be easily computed using similar arguments in \citet{ Wood04, XH12}, it can also be optimized using the Newton-Raphson algorithm with the same computational costs. 

\subsection{The Newton-Raphson Implementation}
In some applications, not only the penalty parameter $\lambda$ in~\eqref{pls1} needs to be carefully selected, it is also important to choose other tuning parameters in the kernel function. For example, the bandwidth parameter $\phi$ in the Gaussian kernel $K(\bm x,\bm z)=\exp(-\|\bm x-\bm z\|_2^2/\phi)$ also plays an important role in the performance of the KRR, as we will illustrate in the Million Song Dataset in Section~\ref{sec:rea}. In such cases, $\dGCV$ can serve as a tool to choose the optimal tuning parameters $\bm{\theta}$ in the kernel function, as long as conditions C1-C4 in Section~\ref{sec:opt} are satisfied. One remaining practical issue is that when the dimension of $\btheta$ is high,  the grid search method for the optimal combination of $\lambda$ and $\btheta$ using $\dGCV$ is no longer feasible. Therefore, it is necessary to develop more efficient algorithms such as the Newton-Raphson type algorithm.

Following \cite{Wood04}, denote $\eta=\log\lambda$ and $\dGCV(\eta,\btheta)=\alpha(\eta,\btheta)/\gamma(\eta,\btheta)$, where 
\bse
\alpha(\eta,\btheta)&=&\frac{1}{N}\Yv^T\{\bI-\bar{\bA}_m(\eta,\btheta)\}^T\bW\{\bI-\bar{\bA}_m(\lambda,\btheta)\}\Yv,\\
\gamma(\eta,\btheta)&=&\left[1-\frac{1}{Nm}\sum_{k=1}^m\tr\{\bA_{kk}(\eta,\btheta)\bW_k\}\right]^2,
\ese 
with $\bar{\bA}_m(\eta,\btheta)$ and $\bA_{kk}(\eta,\btheta)$'s defined in~\eqref{Am}. Then the first and second partial derivatives of $\log\left[\dGCV(\eta,\btheta)\right]$ can be straightforwardly obtained as
\[
\frac{\partial \log\left[\dGCV(\eta,\btheta)\right]}{\partial \vartheta}=\frac{1}{\alpha(\eta,\btheta)}\frac{\partial \alpha(\eta,\btheta)}{\partial\vartheta}-\frac{1}{\gamma(\eta,\btheta)}\frac{\partial \gamma(\eta,\btheta)}{\partial\vartheta},\qquad \vartheta=\eta\text{ or }\btheta.
\]
\[
\begin{split}
\frac{\partial^2 \log\left[\dGCV(\eta,\btheta)\right]}{\partial \vartheta\partial\varrho^T}&=-\frac{1}{\alpha^2(\eta,\btheta)}\left[\frac{\partial \alpha(\eta,\btheta)}{\partial\vartheta}\right]\left[\frac{\partial \alpha(\eta,\btheta)}{\partial\varrho}\right]^T+\frac{1}{\alpha(\eta,\btheta)}\frac{\partial^2 \alpha(\eta,\btheta)}{\partial\vartheta\partial\varrho^T}\\
&+\frac{1}{\gamma^2(\eta,\btheta)}\left[\frac{\partial \gamma(\eta,\btheta)}{\partial\vartheta}\right]\left[\frac{\partial \gamma(\eta,\btheta)}{\partial\varrho}\right]^T-\frac{1}{\gamma(\eta,\btheta)}\frac{\partial^2 \gamma(\eta,\btheta)}{\partial\vartheta\partial\varrho^T},\qquad \vartheta,\varrho=\eta\text{ or }\btheta.
\end{split}
\]
By definitions of $\alpha(\eta,\btheta)$ and $\gamma(\eta,\btheta)$, straightforward matrix calculus yields that it remains to compute partial derivatives of $\bA_{kl}(\eta,\btheta)=\bK_{kl}^T(\btheta)\left[\bK_{kk}(\btheta)+n_ke^{\eta}\bI_k\right]^{-1}$ with $\bK_{kl}=\left[K(\bm x_{i},\bm x_{j};\btheta)\right]_{i\in S_k,j\in S_l}$ for $l,k=1,\dots,m$. It is straightforward to show that
\bse
&&\hskip-2em\frac{\partial\bA_{kl}(\eta,\btheta)}{\partial\eta}=-n_ke^{\eta} \bK_{kl}^T(\btheta)\bK_{kk}^{\ddag2},\qquad \frac{\partial\bA_{kl}(\eta,\btheta)}{\partial\theta_c}= \frac{\partial\bK_{kl}^T(\btheta)}{\partial\theta_c}\bK_{kk}^{\ddag}-\bK_{kl}^T(\btheta)\bK_{kk}^{\ddag}\frac{\partial\bK_{kk}(\btheta)}{\partial\theta_c}\bK_{kk}^{\ddag},\\
&&\hskip-2em\frac{\partial^2\bA_{kl}(\eta,\btheta)}{\partial\eta^2}=-n_ke^{\eta} \bK_{kl}^T(\btheta)\bK_{kk}^{\ddag2}+2n_k^2e^{2\eta} \bK_{kl}^T(\btheta)\bK_{kk}^{\ddag3},\\
&&\hskip-2em\frac{\partial^2\bA_{kl}(\eta,\btheta)}{\partial\eta\partial\theta_c}= -n_ke^{\eta}\left\{\frac{\partial\bK_{kl}^T(\btheta)}{\partial\theta_c}-\bK_{kl}^T(\btheta)\bK_{kk}^{\ddag}\frac{\partial\bK_{kk}(\btheta)}{\partial\theta_c}\right\}\bK_{kk}^{\ddag2}+n_ke^{\eta}\bK_{kl}^T(\btheta)\bK_{kk}^{\ddag2}\frac{\partial\bK_{kk}(\btheta)}{\partial\theta_c}\bK_{kk}^{\ddag},\\
&&\hskip-2em\frac{\partial^2\bA_{kl}(\eta,\btheta)}{\partial\theta_{c_1}\partial\theta_{c_2}}= \frac{\partial^2\bK_{kl}^T(\btheta)}{\partial\theta_{c_1}\partial\theta_{c_2}}\bK_{kk}^{\ddag}-\frac{\partial\bK_{kl}^T(\btheta)}{\partial\theta_{c_1}}\bK_{kk}^{\ddag}\frac{\partial\bK_{kk}^T(\btheta)}{\partial\theta_{c_2}}\bK_{kk}^{\ddag}-\frac{\partial\bK_{kl}^T}{\partial\theta_{c_2}}(\btheta)\bK_{kk}^{\ddag}\frac{\partial\bK_{kk}(\btheta)}{\partial\theta_{c_1}}\bK_{kk}^{\ddag}\\&&\hskip5em+\bK_{kl}^T(\btheta)\bK_{kk}^{\ddag}\frac{\partial\bK_{kk}(\btheta)}{\partial\theta_{c_2}}\bK_{kk}^{\ddag}\frac{\partial\bK_{kk}(\btheta)}{\partial\theta_{c_1}}\bK_{kk}^{\ddag}-\bK_{kl}^T(\btheta)\bK_{kk}^{\ddag}\frac{\partial^2\bK_{kk}(\btheta)}{\partial\theta_{c_1}\partial\theta_{c_2}}\bK_{kk}^{\ddag}\\&&\hskip5em+\bK_{kl}^T(\btheta)\bK_{kk}^{\ddag}\frac{\partial\bK_{kk}(\btheta)}{\partial\theta_{c_1}}\bK_{kk}^{\ddag}\frac{\partial\bK_{kk}(\btheta)}{\partial\theta_{c_2}}\bK_{kk}^{\ddag},\quad \text{ for } \btheta=(\theta_1,\cdots,\theta_D), c,c_1,c_2=1,\cdots,D,
\ese
where $\bK_{kk}^{\ddag}=\left[\bK_{kk}(\btheta)+n_ke^{\eta}\bI_k\right]^{-1}$, $k,l=1,\cdots,m$ and all matrix derivatives are taken element-wise.

 It is straightforward to show that the computational complexity of first and second derivatives of $\log\left[\dGCV(\eta,\btheta)\right]$ are the same as that of $\dGCV$, which makes the Newton-Raphson type algorithm feasible. However, it is worth pointing out that $\log\left[\dGCV(\eta,\btheta)\right]$ is not a convex function of $\eta$ and $\btheta$, hence there is no guarantee that a Newton-Raphson type algorithm will converge to the global minimizer. Numerical suggestions such as those in~\cite{Wood04} may be useful for developing more efficient algorithms, which will be an interesting further research topic.

\section{Asymptotic Properties}\label{sec:thry}

In this section, we will show that the proposed dGCV criterion in (\ref{gcv-m}) is ``globally optimal" under some conditions. We first introduce some notation. Denote $\Px$, $\Peps$, $\Pepsx$ as the probability measures of covariate $X$, error process $\varepsilon$ and their joint probability measure. Similarly, $\EE$ and $\Vare$ denote the expectation and variance under the probability measure $\Peps$. Let $\lambda_{\max}(\bA)$ and $\sigma_{\max}(\bA)$ and $\tr(\bA)$ be the largest eigenvalue and the largest singular value of the matrix $\bA$, respectively.  We use $\xrightarrow{\Pb}$ to denote the convergence in probability measure $\Pb$ and $O_{\Pb}(\cdot)$, $o_{\Pb}(\cdot)$ as defined in the conventional way. For any function $f(x): \mathcal{X}\to \mathbb{R}$, let $\|f\|_{\sup}=\sup_{x\in \mathcal{X}}|f(x)|$ and $\Pb f=\int_{\mathcal X}f(x)\,d\Pb.$ Finally, let $\Pb_n$ denote the empirical probability measure based on i.i.d samples of size $n$ from the probability measure $\Pb$. 
\subsection{Asymptotic Optimality of dGCV }\label{sec:opt}

The following regularity conditions are needed to show the optimality of $\dGCV$.
\begin{description}
	\item[[C1\!\!\!]]$\frac{1}{m}\sum_{l=1}^m\lambda_{\max}\left\{(\bK_{ll}+n_l\lambda\bI_l)^{-2}\left(\frac{1}{m}\sum_{k=1}^m\bK_{kl}^T\bK_{kl}\right)\right\}=O_{\Px}(1)$;
	\item[[C2\!\!\!]] $N\bar{R}(\lambda|\Xv)\xrightarrow{\Px} \infty$ as $N\to\infty$;
	\item[[C3\!\!\!]] (a) The weights $w_i$'s are nonnegative such that $\sum_{i=1}^{N}w_i=N$ and that $\max_{1\leq i\leq N}w_i\leq W$ for some constant $W>0$; (b) $\frac{1}{Nm}\sum_{k=1}^m \tr\{\bA_{kk}(\lambda)\}=o_{\Px}(1)$ as $N\to\infty$.
	\item[[C4\!\!\!]] $\frac{[N^{-1}\tr\{\bar\bA_m(\lambda)\bW\}]^2}{\left[N^{-1}\tr\{\bar\bA_m^T(\lambda)\bW\bar\bA_m(\lambda)\}\right]}=o_{\Px}(1)$ as $N\to\infty$.
\end{description}

Intuitively, condition C1 requires that some similarities among sub-data sets. If all $\bK_{kl}$'s are similar to $\bK_{ll}$, we can expect $\lambda_{\max}\left\{(\bK_{ll}+n_l\lambda\bI_l)^{-2}\left(\frac{1}{m}\sum_{k=1}^m\bK_{kl}^T\bK_{kl}\right)\right\}\leq 1$, in which case C1 holds. {In Section~\ref{lowcon}, we shall show that one sufficient condition for C1 to hold is to ensure that the ``maximal marginal degrees of freedom" \citep{Bach13} $d_{\lambda}$ defined in~\eqref{mdf} is sufficiently small compared to $N/m$.} Condition C2 is a widely used condition to ensure the optimality of the GCV to hold, for example, see \citet{CW79,Li86,GM05,XH12}. It is a mild condition for nonparametric regression problems, where the parametric rate $O(N^{-1})$ is unattainable for the estimation risk. For example, for kernel ridge regression models with polynomially or exponentially decaying kernel functions, condition C2 holds \citep{ZDW15}.  However, it does raise a flag for the application of the dGCV when a finite rank kernel is used, in which case the optimal rate of $\bar{R}(\lambda|\Xv)$ is of the order $O(N^{-1})$ \citep{ZDW15}. Nevertheless, without condition C2, it is questionable whether there exists an asymptotically optimal selection procedure for the tuning parameter $\lambda$ \citep{Li86}.  
{
	\begin{remark}
		\label{rem1}
		Condition C3(a) has an important implication for the $\dGCV^*(\lambda)$ defined in Section~\ref{sec:cc}. When leaving out a portion of data as suggested in Section~\ref{sec:cc}, the resulting weights become $w_i=N/(\sum_{k=1}^{m^*}n_k)$ if $i\in\cup_{k=1}^{m^*}S_k$ and $w_i=0$ otherwise. Condition C3(a) requires that the number of data points remained (i.e., $\sum_{k=1}^{m^*}n_k$) must be of the same order as $N$. Therefore, more data points need to be retained as the sample size $N$ grows. Furthermore, when all sub-datasets under the divide-and-conquer procedure are roughly of the same size, Condition C3(a) essentially requires that $m^*/m=c$ for some absolute constant $0<c\leq 1$. From the computational point of view, it is worth to use a $m^*<m$ only when $N>>m^2$. Therefore, a general rule of thumb for the choice of $m^*$ is that it should only be used when  $N>>m^2$ and if used it cannot be too small compared to $m$.
	\end{remark}	
}

It turns out that, under conditions C1-C2 and C3(a), $\bar{U}(\lambda|\Xv)$ defined in~(\ref{U}) is ``globally optimal." 
\begin{lemma}
	\label{lem1}
	Under Conditions C1--C2 and C3(a), for a fixed $\lambda$, we have that
	\be
	\label{eq:lem1}
	\bar{U}(\lambda|\Xv)-\bar{L}(\lambda|\Xv)-\frac{1}{N}\beps^T\bW\beps=o_{\Pepsx}\{\bar{L}(\lambda|\Xv)\}.
	\ee
\end{lemma}
The proof is given in the Appendix.

Lemma~\ref{lem1} states that when $\sigma^2$ is known, minimizing $\bar{U}(\lambda|\Xv)$ with respect to $\lambda$ is asymptotically equivalent to minimizing the empirical true loss function  $\bar{L}(\lambda|\Xv)$. However, it is rarely the case that one has complete knowledge of $\sigma^2$. In this sense, the proposed dGCV is more practical and it can be shown to be ``globally optimal" as well, under some additional conditions. 

\begin{theorem}
	\label{thm1}
	Under Conditions C1--C4, for a fixed $\lambda$, we have that
	\be
	\label{eq:thm1}
	{\rm dGCV}(\lambda|\bX)-\bar{L}(\lambda|\bX)-\frac{1}{N}\beps^T\bW\beps=o_{\Pepsx}\{(\bar{L}(\lambda|\bx)\}.
	\ee
\end{theorem}
The proof is given in the Appendix.

Similar to Lemma~\ref{lem1}, Theorem~\ref{thm1} shows that minimizing $\dGCV(\lambda|\Xv)$ amounts to minimizing the true conditional loss function $\bar{L}(\lambda|\Xv)$, although additional conditions C3(b)-C4 are needed. Condition C3(b) is rather mild in that it essentially requires that the effective degrees of freedom to be negligible compared to the sample size, which is typically true for non-parametric function estimators in most settings of interest. In addition, C3(b) becomes trivial when $m\to\infty$ because by definition we have that $\tr\{\bA_{kk}(\lambda)\}\leq n_k$, $k=1,\dots,m.$ When the entire data set is used at once ($m=1$), condition C4 reduces to the well known condition $[N^{-1}\tr\{\bA(\lambda)\}]^2/[N^{-1}\tr\{\bA^2(\lambda)\}]=o(1)$ in the literature \citep{CW79,Li86,GM05,XH12}. For example, for smoothing splines, we typically have $\tr\{\bA(\lambda)\}=O(\lambda^{-1/s})$ and $\tr\{\bA^2(\lambda)\}\asymp O(\lambda^{-1/s})$ for some $s>1$. Then as long as $\lambda^{-1/s}/N\to 0$, which covers the most region of practical interest for $\lambda$, we have that $[N^{-1}\tr\{\bA(\lambda)\}]^2/[N^{-1}\tr\{\bA^2(\lambda)\}]\to 0$ as $N\to\infty$. Condition C4 can be viewed as an extension of this commonly used condition to the divide-and-conquer regime.

\subsection{Low-level Sufficient Conditions for  C1 and C4}
\label{lowcon}
In this subsection, for simplicity, we only consider uniform weights with $w_1=\dots=w_N=1$ and equal sample sizes $n_1=\cdots=n_m=n$ in this subsection. We first establish a low-level sufficient condition for C1. Following \cite{Bach13}, define the ``maximal marginal degrees of freedom" as 
\be
\label{mdf}
d_{\lambda}=N\|\diag\{\bK(\bK+N\lambda\bI_N)^{-1}\}\|_{\infty},
\ee
where $\|\cdot\|_{\infty}$ stands for the matrix infinity norm. Note that $\bA(\lambda)=\bK(\bK+N\lambda\bI_N)^{-1}\}$ is the hat matrix~\eqref{Am} with $m=1$ and ${\rm df}_{\lambda}=\tr\left[\bA(\lambda)\right]=\|\diag\{\bK(\bK+N\lambda\bI_N)^{-1}\}\|_{1}$ defines the ``effective degrees of freedom" \citep{Gu02} for the KRR using the entire dataset at once. In this sense, the ``maximal marginal degrees of freedom" $d_{\lambda}$ provides an upper bound for the ``effective degree of freedom" ${\rm df}_{\lambda}$ due to the inequality ${\rm df}_{\lambda}\leq d_{\lambda}$, and hence gives another measure for the model complexity.
\begin{description}
	\item[[C1'\!\!\!]]Let $r={\rm rank}(\bK)$ and $d_{\lambda}$ be the ``maximal marginal degrees of freedom" defined in~\eqref{mdf}, we assume that
	\be
	\frac{md_{\lambda}\left(\log r+\log m\right)}{N}=o_{\mathbb{P}_X}(1),
	\ee
	as $N\to\infty$ for either a finite $m$ or $m\to\infty.$
\end{description}
Condition C1' ensures that the number of partitions $m$ cannot be too large compared to the total sample size $N$, depending on the magnitude of $d_{\lambda}$, which is consistent with findings in the literature \citep{ZDW15, SC17}. With a large $m$, condition C1' maybe violated if there is a significant number of outliers, leading to a potentially large $d_{\lambda}$.

\begin{lemma}
	\label{lem3}
	Condition C1' is sufficient for condition C1.
\end{lemma}
The proof is given in the Appendix.

 Next we proceed to derive sufficient conditions for condition C4. When the entire data set is used at once ($m=1$) and conditional on observed covariate $\bX$, condition C4 reduces to the well known condition $[N^{-1}\tr\{\bA(\lambda)\}]^2/[N^{-1}\tr\{\bA^2(\lambda)\}]=o(1)$ in the literature \citep{CW79,Li86,GM05,XH12}. For example, for smoothing splines, we typically have $\tr\{\bA(\lambda)\}=O(\lambda^{-1/s})$ and $\tr\{\bA^2(\lambda)\}\asymp O(\lambda^{-1/s})$ for some $s>1$. In this case, as long as $\lambda^{-1/s}/N\to 0$, which covers the most region of practical interest for $\lambda$, we have that $[N^{-1}\tr\{\bA(\lambda)\}]^2/[N^{-1}\tr\{\bA^2(\lambda)\}]\to 0$ as $N\to\infty$. Condition C4 can be viewed as an extension of this commonly used condition to the divide-and-conquer regime, whose justification, however, is much less straightforward. 

We first provide some heuristic insights behind our proof. Define
\be
\label{Q}
Q(\lambda|\Xv)=\int_{\mathcal{X}}\Vare\{\bar{f}(x)\}^2\,d\Px(x)	=\frac{1}{m^2}\sum_{k=1}^m\int_{\mathcal{X}}\Vare\{\widehat{f}_k(x)\}\,d\Px(x).
\ee
Let $\ePx_{X,N}$ be the empirical measure based on sample $\{X_1,\dots,X_N\}$, and $\ePx_{X,n_{k}}$ be the empirical measure based on the $k$-th sub-sample $\{X_i\}_{i\in S_k}$. It is straightforward to show that
\be
\label{Q1}
Q_1(\lambda|\Xv)=\sigma^2\frac{\tr\{\bar{\bA}_m^T(\lambda)\bar{\bA}_m(\lambda)\}}{N}=\int_{\mathcal{X}}\Vare\left\{\bar{f}(x)\right\}^2\,d\ePx_{X,N}(x),
\ee
\be
\label{Q2}
Q_2(\lambda|\Xv)=\sigma^2\frac{1}{Nm}\sum_{k=1}^m\tr\{\bA_{kk}^2(\lambda)\}=\frac{1}{m^2}\sum_{k=1}^m\int_{\mathcal{X}}\Vare\{\widehat{f}_k(x)\}\,d\ePx_{X,n_k}(x).
\ee
Intuitively, $Q_1(\lambda|\Xv)$ and $Q_2(\lambda|\Xv)$ are two empirical versions of $Q(\lambda|\Xv)$ and should be close to each other. The formal proof utilizes the uniform ratio limit theorems for empirical processes \citep{P95} to show  $Q_1(\lambda|\Xv)/Q(\lambda|\Xv)=1+o_{\Px}(1)$ and $Q_2(\lambda|\Xv)/Q(\lambda|\Xv)=1+o_{\Px}(1)$, then with the help of condition C4'(a), we can show condition C4 holds.  

Let $\mathcal{N}(\epsilon,\|\cdot\|_{\Pb_{X,n}},\mathcal{F})$
be the $\epsilon$-covering number \citep{P86} of a function class $\mathcal{F}$
with the empirical norm $\|f\|_{\Pb_{X,n}}=\sqrt{n^{-1}\sum_{i=1}^{n}f^2(X_i)}$. 
Following conditions are sufficient to ensure condition C4. 

\begin{enumerate}
	\item[[C4'\!\!\!]](a) $\frac{1}{m}\sum_{k=1}^{m}\left[\frac{1}{N}\tr\{\bA_{kk}(\lambda)\}\right]^2/\left[\frac{1}{N}\tr\{\bA_{kk}^2(\lambda)\}\right]=o_{\Px}(1)$;
	\item[[C4'\!\!\!]](b) There exists a positive sequence $\{V_n\}$ such that as $V_n\to 0$, it holds that $V_n\left[\frac{1}{m}\sum_{k=1}^{m}\int_{\mathcal{X}}\Vare\{\widehat{f}_k(x)\}\,d\Px(x)\right]^{-1}=O_{\Px}(1)$, $\max_{1\leq k\le m}\|\Vare\{\widehat{f}_k(x)\}\|_{\sup}=O_{\Px}(V_n)$ and $nV_n\to\infty$ as $n\to\infty$;
	\item[[C4'\!\!\!]](c) There exists a sequence $\{H_n\}$ such that $H_n\left[\frac{n}{m}\sum_{k=1}^{m}\int_{\mathcal{X}}\Vare\{\widehat{f}_k(x)\}\,d\Px(x)\right]^{-1}=O_{\Px}(1)$, $\max_{1\leq k\le m}[\int_{\mathcal{X}}\Vare\{\widehat{f}_k'(x)\}\,d\Px(x)/\int_{\mathcal{X}}\Vare\{\widehat{f}_k(x)\}\,d\Px(x)]=O_{\Px}(H_n^2)$,  and $nH_nV_n-(\log m)^2\to\infty$ as $n\to\infty$. Here, $\widehat{f}_k'(x)$ denotes the derivative of $\widehat{f}_k(x)$;
	\item[[C4'\!\!\!]](d) For the function class $\mathcal{F}_0=\{f: \|f\|_{\sup}\leq 1, J_1(f)=\int_{\mathcal{X}}\left\{f'(x)\right\}^2\,d\Px(x)\leq 1\}$, we have that $\mathcal{N}(\epsilon,\|\cdot\|_{\Pb_{X,n}},\mathcal{F}_0)\leq \exp(C_0/\epsilon)$ for some constant $C_0>0$ with probability approaching one as $n\to\infty$.
\end{enumerate}

\begin{lemma}
	\label{lem2}
	For a tuning parameter $\lambda$ satisfying conditions C4'(a)-(d), one has that
	\[
	\left\{\frac{1}{N}\tr(\bar\bA_m)\right\}^2/\left\{\frac{1}{N}\tr(\bar\bA_m^T\bar\bA_m)\right\}=o_{\Px}(1).
	\] 
\end{lemma}
The proof is given in the Appendix.

Condition C4'(a) is a mild condition as we have discussed at the beginning of this subsection. Condition C4'(b) essentially states that the supremum norm and the $L_1$ norm of the variance function $\Vare\{\widehat{f}_k(x)\}$ are of the same order, which is reasonable when all $\Vare\{\widehat{f}_k(x)\}$'s similarly well-behaved  within the support of covariate $X$. In addition, we should restrict our attention to the range of $\lambda$ such that $n\Vare\{\widehat{f}_k(x)\}\to \infty$, $k=1,\dots,m$. Recall the discussion in subsection~\ref{locvsglob}, the optimal $\bar{f}$ can only be obtained when the risk (\ref{riskk}) is dominated by the variance term  $\Vare\{\widehat{f}_k(x)\}$ for each individual $\widehat{f}_k(x)$. Hence, letting $nV_n\to\infty$ is reasonable based on the condition C2. Condition C4'(c) essentially asserts that $H_n$ and $n V_n$ are of the same order. For the smoothing spline case, the derivative $\widehat{f}_k'$ is typically more variable than $\widehat{f}_k$ such that one can expect $H_n\to\infty$. For example, \cite{RR83} gives the exact rates of convergence for cubic smoothing spline, that is $\int_{\mathcal{X}}\Vare\{\widehat{f}_k(x)\}\,d\Px(x)\asymp n^{-1}\lambda^{-1/4}$, $\int_{\mathcal{X}}\Vare\{\widehat{f}_k'(x)\}\,d\Px(x)\asymp n^{-1}\lambda^{-3/4}$. In this case, we have that $H_n\asymp \lambda^{-1/4}$ and $nV_n\asymp\lambda^{-1/4}$. A thorough theoretical investigation of $H_n$ and $V_n$ is difficult in general, though our simulation study (unreported) suggests condition C4'(c) to be reasonable for many reproducing kernels.

Finally, condition C4'(d) holds when the empirical measure $\Pb_{X,n}$ is replaced by $\Px$, see, e.g., \cite{VG00}. One can generally expect it to hold when the sample size $n$ is large. The upper bound of the random covering number $\mathcal{N}(\epsilon,\|\cdot\|_{\Pb_{X,n}},\mathcal{F}_0)$ determines the rate of convergence of the empirical processes $Q_1(\lambda|\Xv)$ and $Q_2(\lambda|\Xv)$ to $Q(\lambda|\Xv)$. And it can be relaxed similarly as given in Theorem~2.1 of \cite{P86}.

{
	\begin{remark}
		One benefit of using high level conditions such as C1, C2 and C4 is that they do not involve the response variable and can be computed efficiently using sample data. To deal with the randomness in covariate X, one can bootstrap/resample/subsample from the observed data, which is especially suitable when the sample size under consideration is extremely large. Through this resampling strategy, one can empirically verify C1, C2 and C4, although rigorous justification of such strategy has not been established and will be an interesting topic for future research. 
	\end{remark}
}
\section{Simulation studies}\label{sec:sim}
In this section, we conduct simulation studies to illustrate the effectiveness of $\dGCV(\lambda)$ in choosing the optimal $\lambda$ for the d-KRR. The data were simulated from the model \be
\label{simmod}
y=2.4\times {\rm beta}(x,30,17)+1.6\times {\rm beta}(x,3,11)
+\varepsilon, \quad x\in [0,1],
\ee
where ${\rm beta}(x,a,b)$ is the density function of the ${\rm Beta}(a,b)$ distribution and $\varepsilon\sim N(0,3^2)$. The covariate $x_i$'s were independently generated from the uniform distribution over the interval $[0,1]$. For each simulation run, we first generated a data set of the size $N=mn$ and then randomly partition the data sets into $m$ sub-data sets of equal sizes. The divide-and-conquer estimator $\bar{f}$ was obtained as given in~(\ref{ave}). 

Let $f^{(\nu)}(\cdot)$ be the $\nu$th derivative of a smooth function $f(\cdot)$. The true function in model~\eqref{simmod} belongs to the Sobolev Hilbert space of $\nu$th order differentiable functions on $[0,1]$ satisfying the periodic boundary conditions $f^{(\nu)}(0)=f^{(\nu)}(1)$ for $\nu=1,\cdots,10$, denoted as $\mathcal{W}_{\nu}(per)$ \citep{Wahba90}. If $\mathcal{W}_{\nu}(per)$ is endowed with the norm $\|f\|_{\mathcal{W}_{\nu}}^2=\left\{\int_{0}^{1}f(x)\,dx\right\}^2+\int_{0}^{1}\{f^{(\nu)}(x)\}^2\,dx$, then it has a reproducing kernel 
\be
\label{SimKern}
K(x,z)=\frac{(-1)^{\nu-1}}{(2\nu)!}B_{2\nu}([x-z]),\quad x,z\in[0,1],
\ee
where $B_{2\nu}(\cdot)$ is the $2\nu$th Bernoulli polynomials \citep{abramowitz1972handbook} and $[x]$ is the fractional part of $x$. In all simulation runs, the tuning parameter $\lambda$ was selected by a grid search for $\log(\lambda)$ over $30$ equally-spaced grid points over the interval $[-10\nu,-5\nu]$. Three approaches were used for the selection of $\lambda$: (i) the distributed GCV (dGCV) approach proposed in~(\ref{gcv-m}); (ii) (ii) the naive GCV (nGCV) approach where a $\wh\lambda_k$ is selected for each individual $\widehat{f}_k$ by minimizing the sub-GCV score ${\rm GCV}_k(\lambda)$ defined in \eqref{gcv1} for $k=1,\cdots,m$ and then the final estimator is obtained by averaging all $\widehat{f}_k$'s; and (iii) the true empirical loss function (TrueLoss) $\bar{L}(\lambda|\Xv)$ defined in (\ref{loss}). The last approach is not practically feasible since it requires the knowledge of the truth $f_0$. It merely serves as the ``golden criterion" to show the effectiveness of other two approaches. For all approaches, we set the weights $w_i=1$ for all $i=1,\dots,N$ and used $\nu=2$ for the kernel~\eqref{SimKern} unless otherwise stated.

\subsection{Performances with Moderate Sample Sizes}
In this subsection, we evaluated performances of the proposed approach with moderate sample sizes  $N=2^i$, $i=8,9,10,11,12$. In this setting, it is still possible to obtain the KRR estimator with the entire data set, i.e., $m=1$, and enables us to evaluate potential loss using the divided-and-conquer approach as opposed to using all data at once.  

\subsubsection{Computational Complexity and Estimation Accuracies}
\label{small}
\begin{figure}[htb!]
	\begin{center}
		\subfigure{\includegraphics[angle=0,width=0.32\textwidth,totalheight=0.3\textwidth]{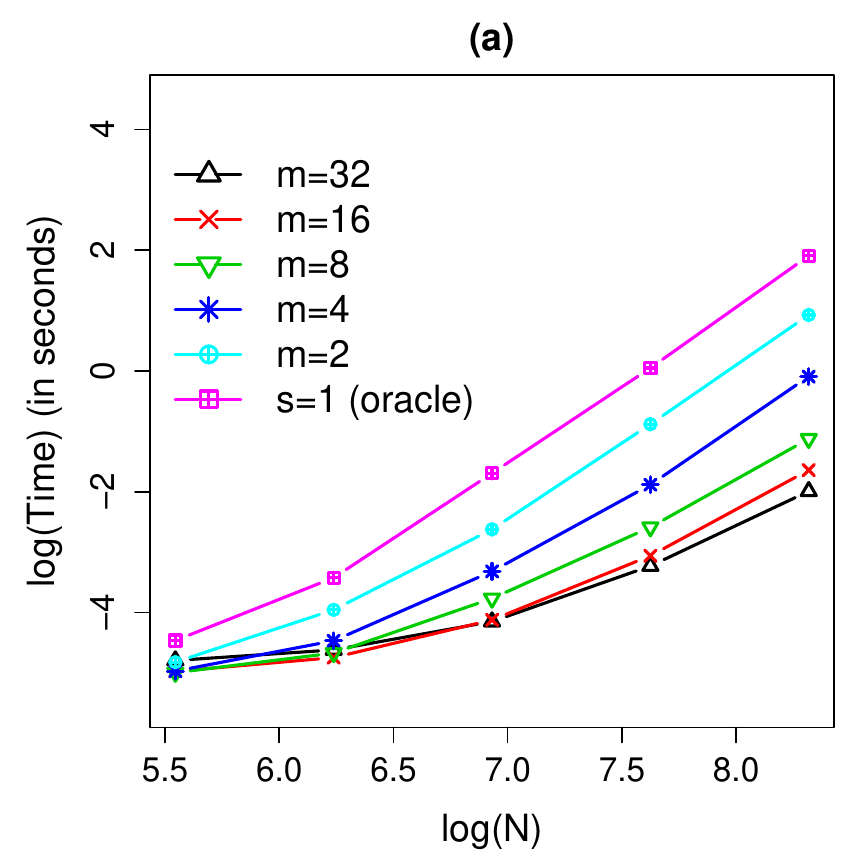}}
		\subfigure{\includegraphics[angle=0,width=0.32\textwidth,totalheight=0.3\textwidth]{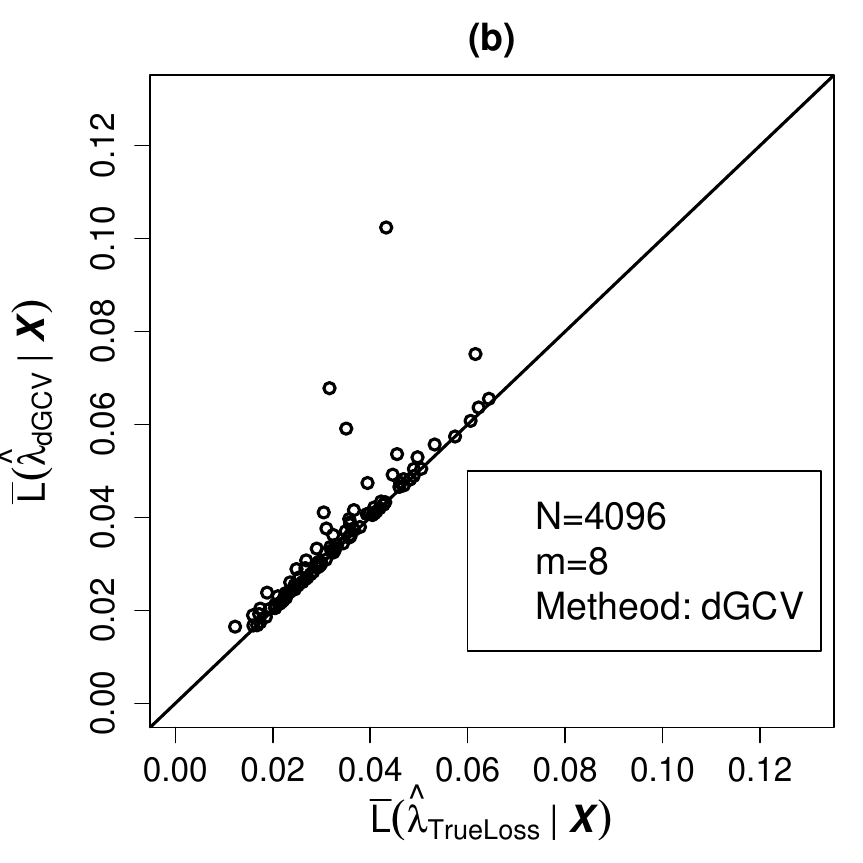}}
		\subfigure{\includegraphics[angle=0,width=0.32\textwidth,totalheight=0.3\textwidth]{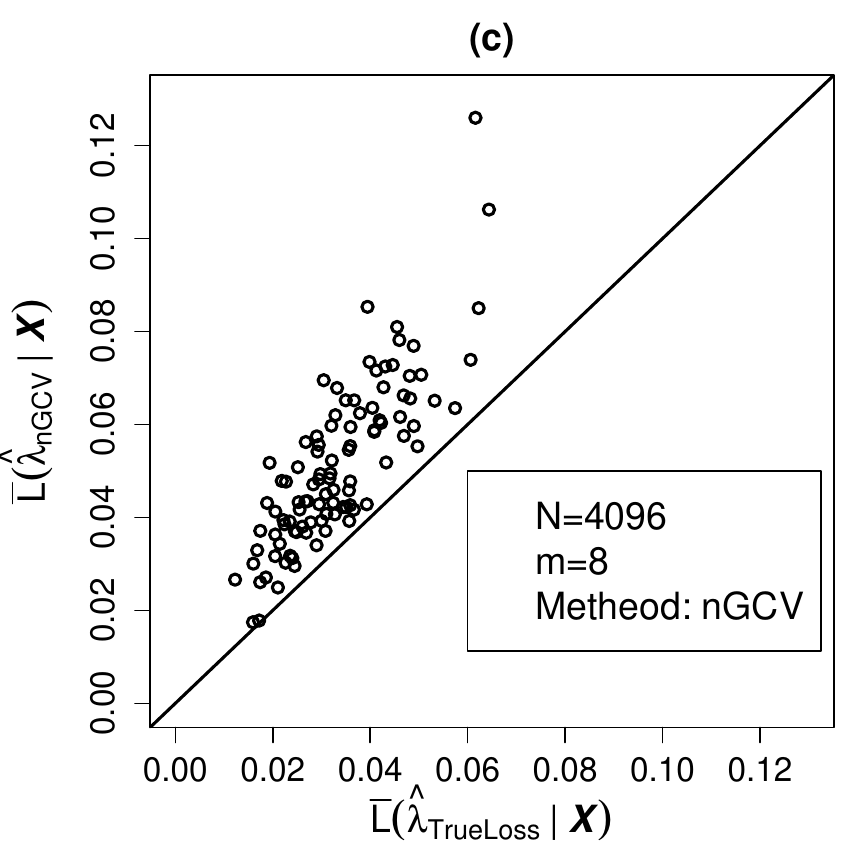}}
		\subfigure{\includegraphics[angle=0,width=0.32\textwidth,totalheight=0.3\textwidth]{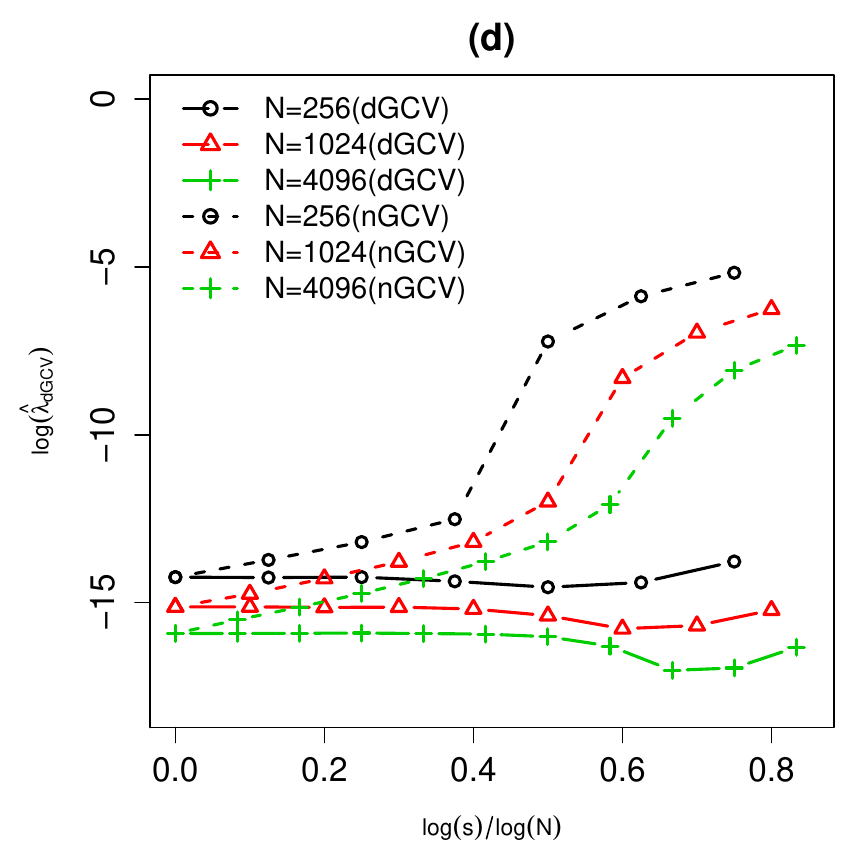}}
		\subfigure{\includegraphics[angle=0,width=0.32\textwidth,totalheight=0.3\textwidth]{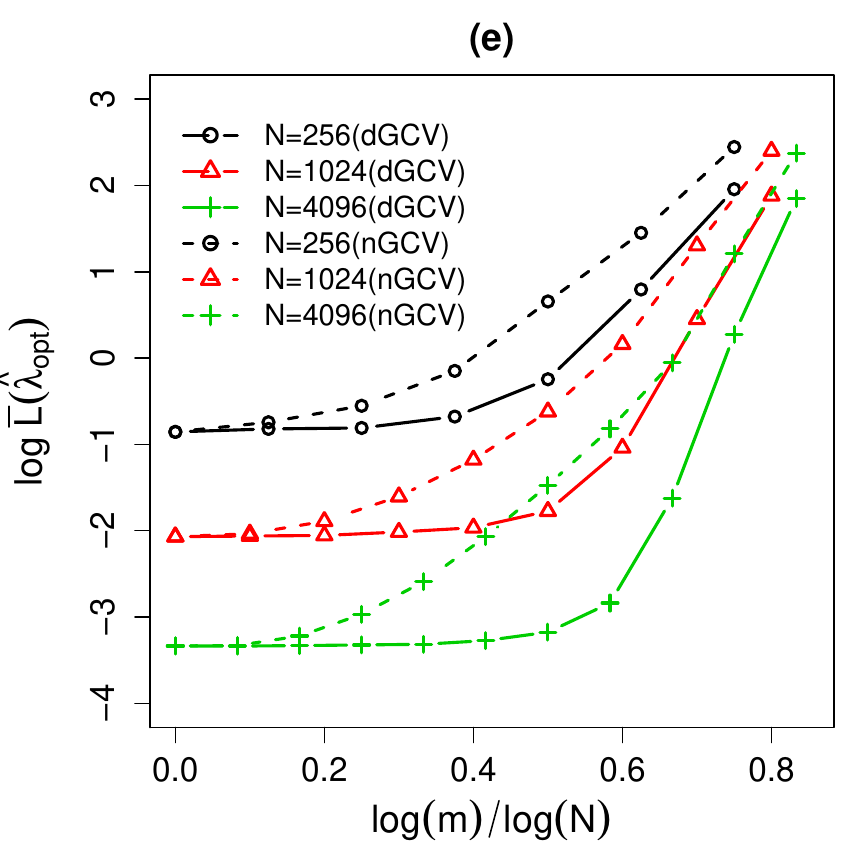}}
		\subfigure{\includegraphics[angle=0,width=0.32\textwidth,totalheight=0.3\textwidth]{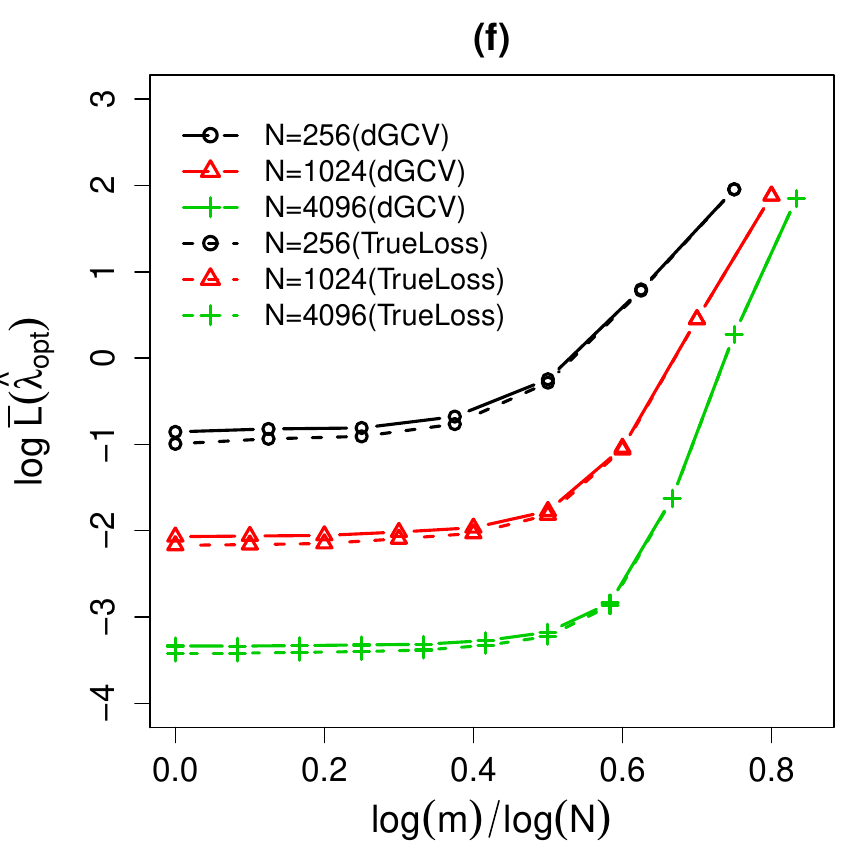}}
		\vspace{-2em}
	\end{center}
	\caption{(a) the logarithm of computational time (in seconds) v.s. $\log(N)$; (b)-(c): scatter plots of true empirical losses of function estimators; (d) the logarithm of averages of selected $\lambda$ v.s. $\log(m)/\log(N)$; (e)-(f): the logarithm of averaged true empirical losses v.s. $\log(m)/\log(N)$. Note that in (d)-(f), $\hat{\lambda}_{\rm opt}$ in the y-axis denotes one of $\hat{\lambda}_{\rm dGCV}$, $\hat{\lambda}_{\rm nGCV}$ and $\hat{\lambda}_{ \rm TrueLoss}$ for each curve. }
	\label{fig-1}
\end{figure}

We first simulate data from model~\eqref{simmod} for various sample sizes $N=2^i$, $i=8,9,10,11,12$ and fit the data with divide-and-conquer regression with  $m=1,2,4,8,16,32$.  Summary statistics based on $100$ simulation runs were illustrated in Figure~\ref{fig-1}(a)-(f).  Figure~\ref{fig-1}(a) illustrates the computational complexity of one evaluation of $\dGCV(\lambda)$ .  All simulation runs were carried out in the software R \citep{Rcite} on a cluster of 100 Linux machines with a total of 100 CPU cores, with each core running at approximately 2 GFLOPS.  We can clearly see that by using the divide-and-conquer strategy, the computational time of the dGCV can be greatly reduced compared to the case when all data were used at once (i.e., $m=1$).

In Figure~\ref{fig-1}(b)-(c), we give some comparisons of the dGCV method and the nGCV method. Figure~\ref{fig-1}(b) shows the scatter plot of true empirical losses, as defined in (\ref{loss}), of the function estimators obtained by minimizing $\dGCV(\lambda)$ versus minimizing the unattainable ``golden criterion" (\ref{loss}) over $100$ simulation runs. As we can see,  majority of points are concentrated around the $45^o$ straight line, which supports our theoretical findings in Theorem~\ref{thm1}. On the contrary, Figure~\ref{fig-1}(c) shows that true empirical losses of the function estimator based on the nGCV approach are generally larger than the minimum possible true losses, indicating that such function estimators are indeed only ``locally" optimal but not ``globally optimal."

In Figure~\ref{fig-1}(d)-(f), we used $N=2^i$ and $m=2^j$ for $j=0,1,\dots, i-2$ and $i=8,10,12$ so that there were at least four data points in each sub-data set. To better understand the differences between the dGCV and the nGCV approaches, Figure~\ref{fig-1}(d) shows how the logarithm of the averages of selected tuning parameters (over $100$ simulation runs), denoted as $\log(\widehat{\lambda}_{opt})$, for each method changes as $m$ increases. As we can see, when $m=1$ they are identical. However, as $m$ increases, the $\lambda$ selected by the nGCV approach consistently increases whereas the $\lambda$ selected by the dGCV method stays about the same until $m$ gets really large and is always smaller than the $\lambda$ selected by the nGCV method. This is consistent with findings in~\cite{ZDW15} where they argue that the locally optimal rate of $\lambda$ for each individual $\widehat{f}_k$ is of the order $O(n^{-4/5})$ with $n=N/m$ whereas the globally optimal rate for $\lambda$ is of the order $O(N^{-4/5})$.

The y-axis of Figure~\ref{fig-1}(e)-(f) is the logarithm of estimation errors $\log \overline{L}(\widehat{\lambda}_{opt})$, where $\overline{L}(\widehat{\lambda}_{opt})$ stands for the averaged true conditional loss defined in~(\ref{loss}) over $100$ simulation runs using different selection approaches for $\lambda$. We can see from Figure~1(e)-(f) that as long as $m$ is not too large compare to $N$, the proposed $\dGCV(\lambda)$ is quite robust in terms of controlling the estimation error as $m$ grows and is almost identical to that of using the true loss function, which is considered as a ``golden criterion." This is consistent with our Theorem~\ref{thm1}. In contrast, estimation errors of the nGCV approach quickly inflates as $m$ increases, which is expected according to our discussion in subsection~\ref{locvsglob}. Finally, it is interesting to point out that as the $\lambda$ selected by the dGCV method starts to drop in Figure~1(d), the estimation errors in Figure~1(e)-(f) start to inflate as well.

\subsubsection{Is It Worth Minimizing $\dGCV(\lambda)$?}
\begin{figure}[htb!]
	\begin{center}
		\subfigure{\includegraphics[angle=0,width=0.42\textwidth,totalheight=0.4\textwidth]{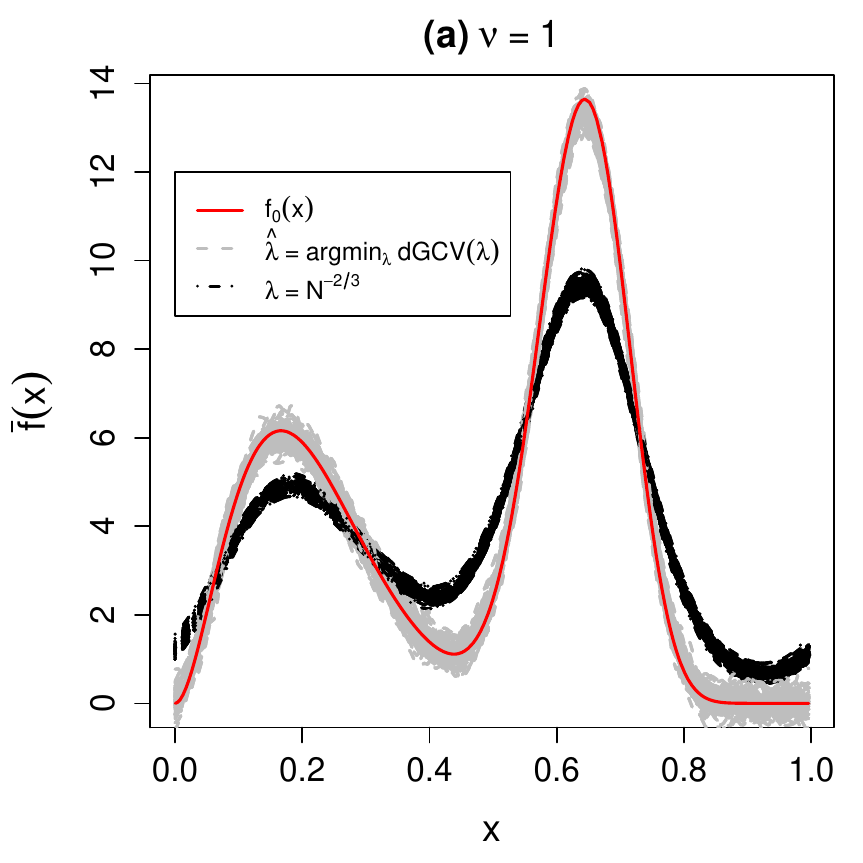}}
		\subfigure{\includegraphics[angle=0,width=0.42\textwidth,totalheight=0.4\textwidth]{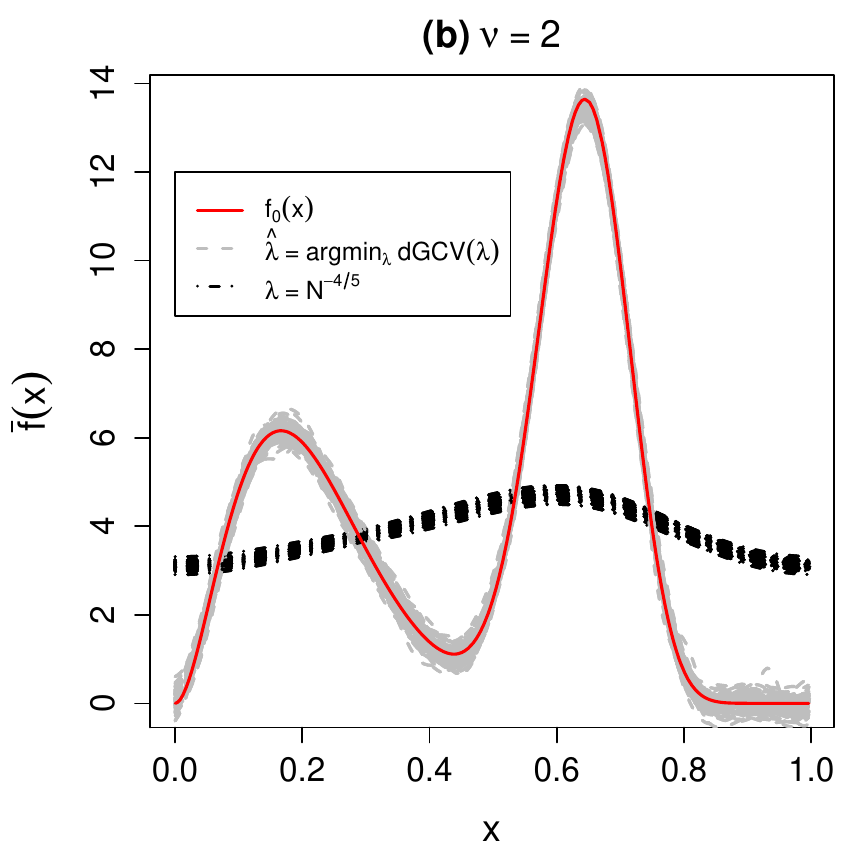}}
		\vspace{-2em}
	\end{center}
	\caption{Estimated functions using Divide-and-conquer KRR with a sample size $N=2^{12}$ and $m=4$. Kernel defined in~\eqref{SimKern} was used with (a) $\nu=1$ and (b) $\nu=2$.}
	\label{fig-3}
\end{figure}

In this subsection, we investigate the issue that whether the extra computational costs in minimizing $\dGCV(\lambda)$ is worthwhile. The optimal rates of $\lambda$ for various reproducing kernels have been well established, see, e.g., \cite{ZDW15}.  In the case of the reproducing kernel~\eqref{SimKern} used in this simulation, the optimal rate for $\lambda$ is of the order $O\left(N^{-\frac{2\nu}{2\nu+1}}\right)$, or in other words, $\lambda_{opt}=CN^{-\frac{2\nu}{2\nu+1}}$ for some constant $C$. One misconception is that the choice of $C$ does not matter much because asymptotically any value of $C$ leads to the same convergence rate for $\overline{f}$. However, for a given sample size, this is far from being true. To illustrate, we fitted the data generated from model~\eqref{simmod} using reproducing kernel~\eqref{SimKern} with $\nu=1$ and $2$, respectively. Resulting function estimators based on $100$ simulation runs with $N=2^{12}=4096$ and $m=4$ were presented in Figure~\ref{fig-3} (a)-(b), where it is apparent that by setting $C=1$, both KRR estimators based on reproducing kernel with $\nu=1$ or $2$ yield much worse estimation accuracies than those of corresponding KRR estimators using $\lambda$ selected by minimizing the proposed $\dGCV(\lambda)$ criterion.   A closer look at the minimization problem~\eqref{pls1}, or equivalently~\eqref{pls2}, suggests that the optimal choice of the constant $C$ in $\lambda_{opt}$ should depend on (a) the magnitude of the kernel function $K(\cdot,\cdot)$; (b) the magnitude of response $\Yv$; (c) the sample size $N$, and therefore can be difficult to obtain in practice. As we have illustrated in Figure~\ref{fig-3}, for a fixed sample size, a carefully chosen constant $C$ (through $\dGCV$ in this case) may have significant impacts on the quality of resulting KRR estimator, for which reason we believe that additional computational costs in minimizing $\dGCV$ is indeed worthwhile.
\subsubsection{The Choice of Number of Partitions $m$}
One remaining issue that we have not addressed theoretically is that how many partitions of data ($m$) should be used in practice for a given sample size $N$. The general guideline for the choice of $m$ is clear: as long as $m$ is not too large compared to $N$, the d-KRR estimator can achieve the optimal convergence rate \citep{ZDW15, SC17}.  However, a practical tool to determine whether $m$ is too large is still lacking. In this subsection, we conducted a simulation study to show that the proposed $\dGCV$ may serve such a purpose.

\begin{figure}[htb!]
	\begin{center}
		\subfigure{\includegraphics[angle=0,width=0.48\textwidth,totalheight=0.4\textwidth]{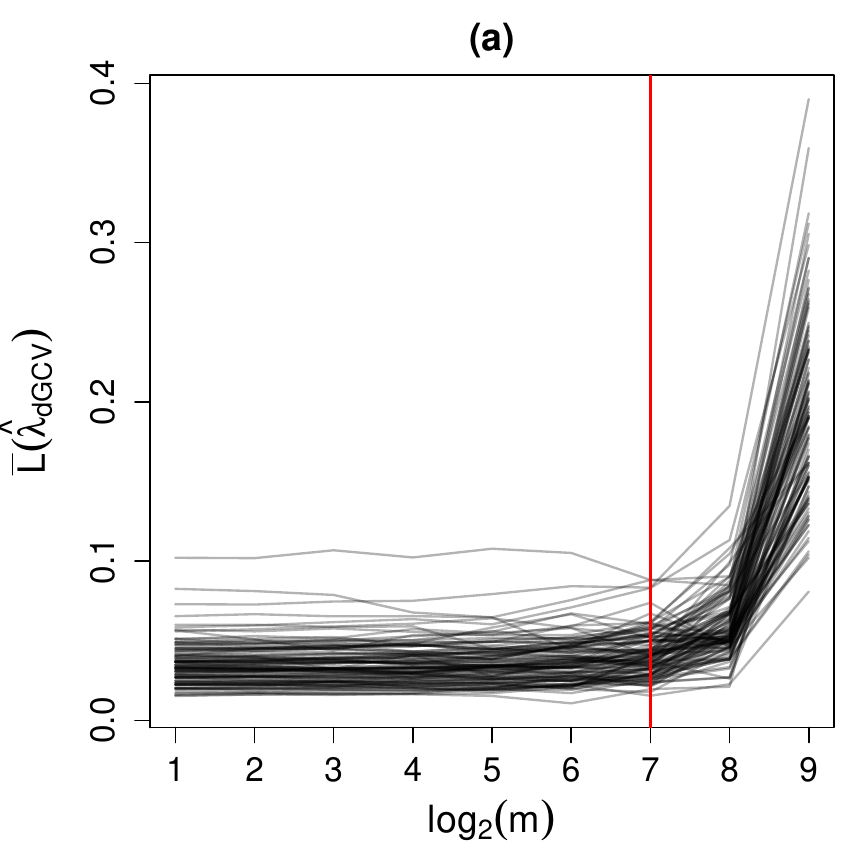}}
		\subfigure{\includegraphics[angle=0,width=0.48\textwidth,totalheight=0.4\textwidth]{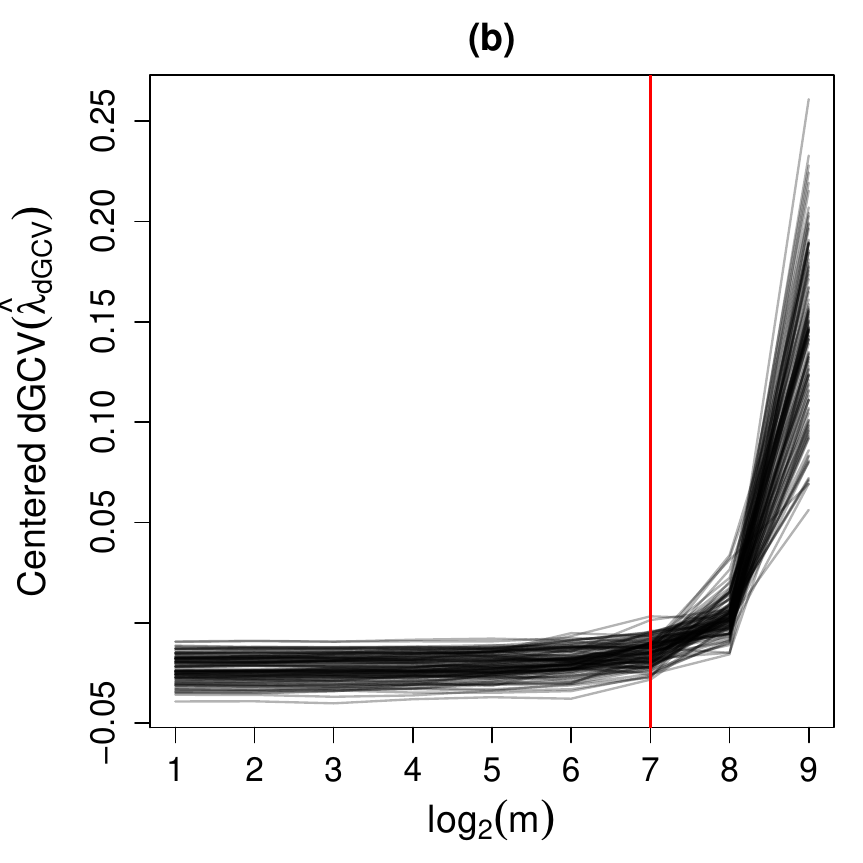}}
		\vspace{-2em}
	\end{center}
	\caption{(a) Empirical true loss defined in~\eqref{loss} using $\lambda$ picked by $\dGCV$ for each $m$; (b) Centered optimal $\dGCV$ score for each $m$; based on 100 simulation runs. ($N=4096$.)}
	\label{fig-4}
\end{figure}

By its definition~\eqref{gcv-m}, $\dGCV(\lambda)$ can also be viewed as a function of $m$, denoted as $\dGCV(\lambda,m)$. Then we can define a profiled version of $\dGCV$ as follows
\be
\label{pgcv}
\dGCV_p(m)=\dGCV(\wh\lambda,m),
\ee
 where $\wh\lambda=\arg\min_{\lambda>0}\dGCV(\lambda,m)$ for a fixed $m$. We simulated data from model~\eqref{simmod} with $N=2^{12}$ for $100$ times and then fitted each data set using d-KRR with $m=2^j$ for $j=1,\cdots, 9$.  Figure~\ref{fig-4}(b) presents  patterns of $100$ centralized version of  $\dGCV_p(m)$, defined as $\dGCV_p(m)-\frac{1}{9}\sum_{j=1}^9\dGCV_p(j)$, as a function of $m$. As comparison, Figure~\ref{fig-4}(a) gives the true empirical loss~\eqref{loss} of each d-KRR estimator using $\wh\lambda=\arg\min_{\lambda>0}\dGCV(\lambda,m)$  for each $m$, where it appears that as long as $m\leq 2^7$, the estimation accuracy of the fitted function remain roughly the same as using the optimal $\lambda$ picked by minimizing  $\dGCV(\lambda,m)$. This coincides with existing theoretical findings in the literature such as \cite{ZDW15} and \cite{SC17}. More importantly, the similarity between Figure~\ref{fig-4} (a) and (b) suggests that the profiled $\dGCV$ score defined in~\eqref{pgcv} can capture the sudden drop in the trajectory of empirical loss as a function of $m$ and therefore determine which $m$ might be too large.  We have tried many other settings and the message remains the same.  This implies that, in practical applications, one can start with a relatively large $m$ and  gradually decrease $m$ until $\dGCV_p(m)$ defined in~\eqref{pgcv} stabilizes. Rigorous justifications of such an approach will be an interesting future research topic.
 
\subsubsection{Performances of $\dGCV$ on Multivariate Functions}
\begin{figure}
	\begin{center}
		\subfigure{\includegraphics[angle=0,width=0.48\textwidth,totalheight=0.4\textwidth]{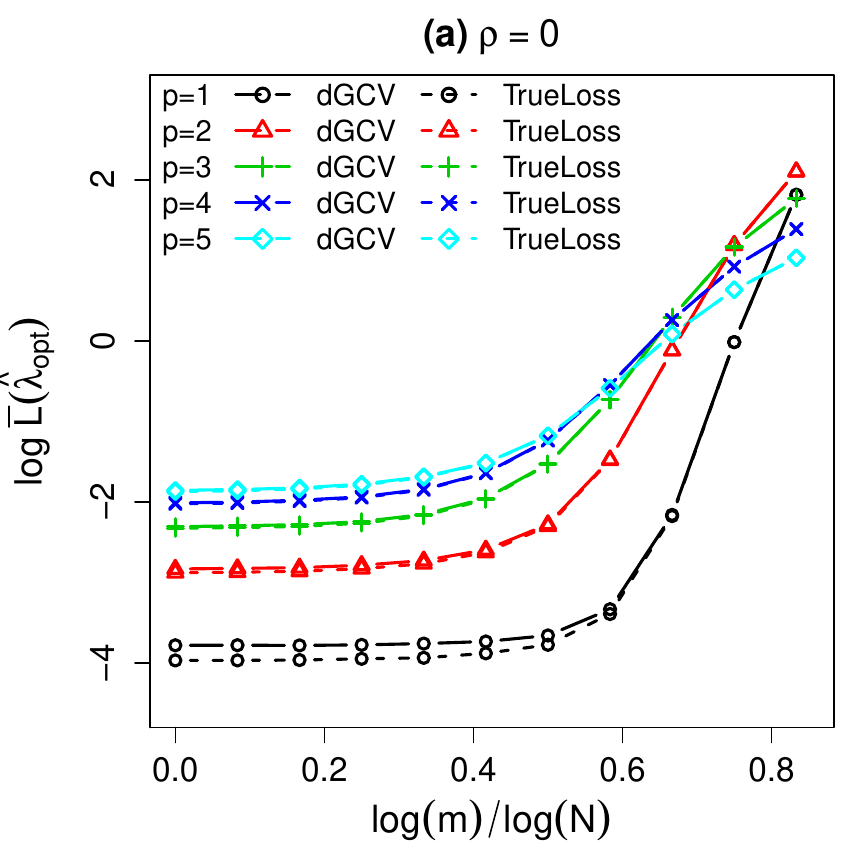}}
		\subfigure{\includegraphics[angle=0,width=0.48\textwidth,totalheight=0.4\textwidth]{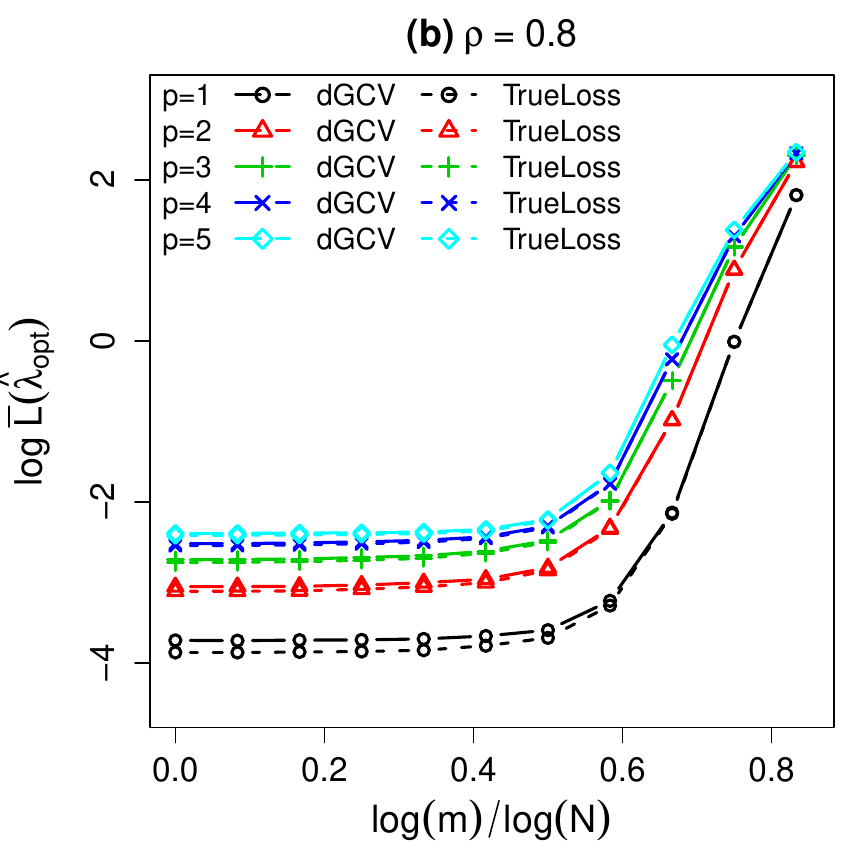}}
		\vspace{-2em}
	\end{center}
	\caption{ The logarithm of averaged true empirical losses v.s. $\log(m)/\log(N)$ with a sample size $N=2^{12}$ and (a) $\rho=0$ (b) $\rho=0.8$.}
	\label{fig-5}
\end{figure}

In this subsection, we investigated the impacts of model dimensionality and correlation among predictors on the performance of $\dGCV$. Let $\bm x=(x_1,\cdots,x_p)^T$, the data was simulated from the following model
\[
y=f(\bm x)=20\left(1-\frac{\|\bm x\|_2}{\sqrt{p}}\right)_{+}^7\left(16\frac{\|\bm x\|_2^2}{p}+7\frac{\|\bm x\|_2}{\sqrt{p}}+1\right)+\varepsilon,\quad\varepsilon\sim N(0,3^2),\quad \bm x\in [0,1]^p,
\]
where $\|\cdot\|_2$ is the Euclidean norm in $\mathbb{R}^p$, function $(r)_{+}=\max(r,0)$ and $x_j$'s are uniformly distributed between $[0,1]$ for $j=1,\cdots,p$. To induce correlations among $x_j$'s,  let $x_j=\Phi(z_j)$ where $(z_1,z_2,\cdots,z_p)^T$ was generated from a $p$-dimensional multivariate normal distribution with mean $0$, variance $1$ and  pairwise correlation coefficient $\rho=0$ or $0.8$.  $f(\bm x)$ is a variate of Wendland's function \citep{schaback2006kernel}. For $p\leq 5$,  we performed the KRR with the reproducing Hilbert kernel space equipped with the kernel 
\[
K(\bm x,\bm z)=\left(1-\frac{\|\bm x-\bm z\|_2}{\sqrt{p}}\right)_{+}^5\left(5\frac{\|\bm x-\bm z\|_2^2}{p}+1\right),\quad \bm x,\bm z\in [0,1]^p,
\]
which is a radial basis function with bounded support for $p\leq 5$, see \cite{schaback2006kernel} for more details. The averaged true empirical losses based on $100$ simulation runs are summarized in Figure~\ref{fig-5}. On one hand, when the dimensionality of $\bm x$ increases from $p=1$ to $5$, the averaged empirical losses gradually increase as expected. However, the averaged empirical losses of d-KRR estimators with $\lambda$ chosen by $\dGCV$ is almost indistinguishable from those of corresponding estimators with $\lambda$ picked by the true empirical loss, regardless of the dimension $p$. This echoes with our theoretical findings in Theorem~\ref{thm1}. On the other hand, as $\rho$ increases from $0$ to $0.8$, the correlations among $x_j$'s seem to have little impact on the estimation accuracies for the estimated overall mean function $f(\bm x)$. In fact, when $\rho=0.8$, the performance of $\dGCV$ is relatively more stable than the case with $\rho=0$ as the dimension $p$ increases. This can be explained by the fact that $f(\bm x)$ only depends on $\|\bm x\|_2$, which is less variable when $p$ increases for the case $\rho=0.8$. For this reason the estimation of $f(\bm x)$ is less affected by the dimensionality when $\rho=0.8$.   

\subsection{Performances with a Large Sample Size}
In this subsection, we investigated two issues when the sample size $N$ is so large that a single machine can no longer handle at once: (a) whether the computational/estimation performance in Section~\ref{small} still persists; (b) what is the impact of the choice of $m^*$ in~\eqref{gcv-m1} on the performance of $\dGCV^*$. 
\subsubsection{Computational Complexity and Estimation Accuracies}
\begin{figure}[htb!]
	\begin{center}
		\subfigure{\includegraphics[angle=0,width=0.32\textwidth,totalheight=0.3\textwidth]{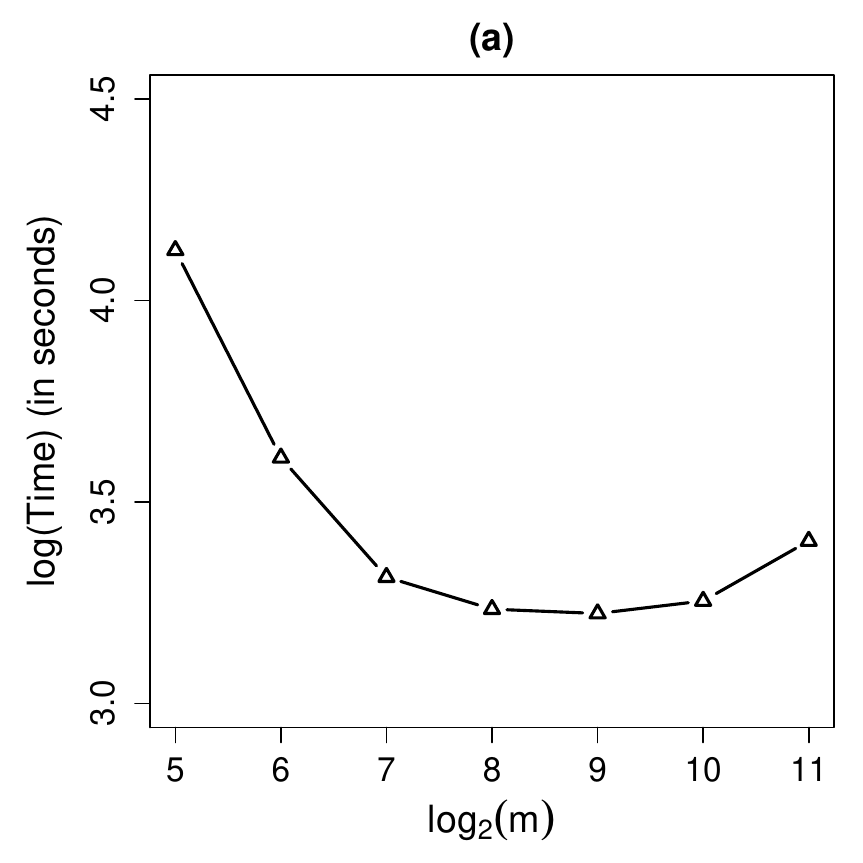}}
		\subfigure{\includegraphics[angle=0,width=0.32\textwidth,totalheight=0.3\textwidth]{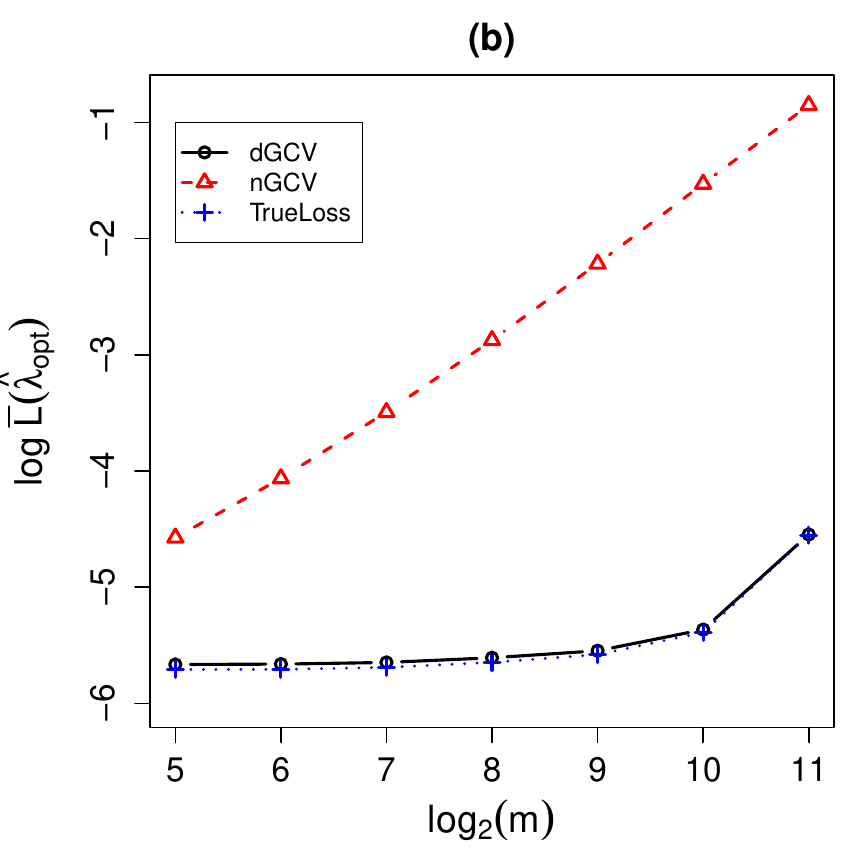}}
		\subfigure{\includegraphics[angle=0,width=0.32\textwidth,totalheight=0.3\textwidth]{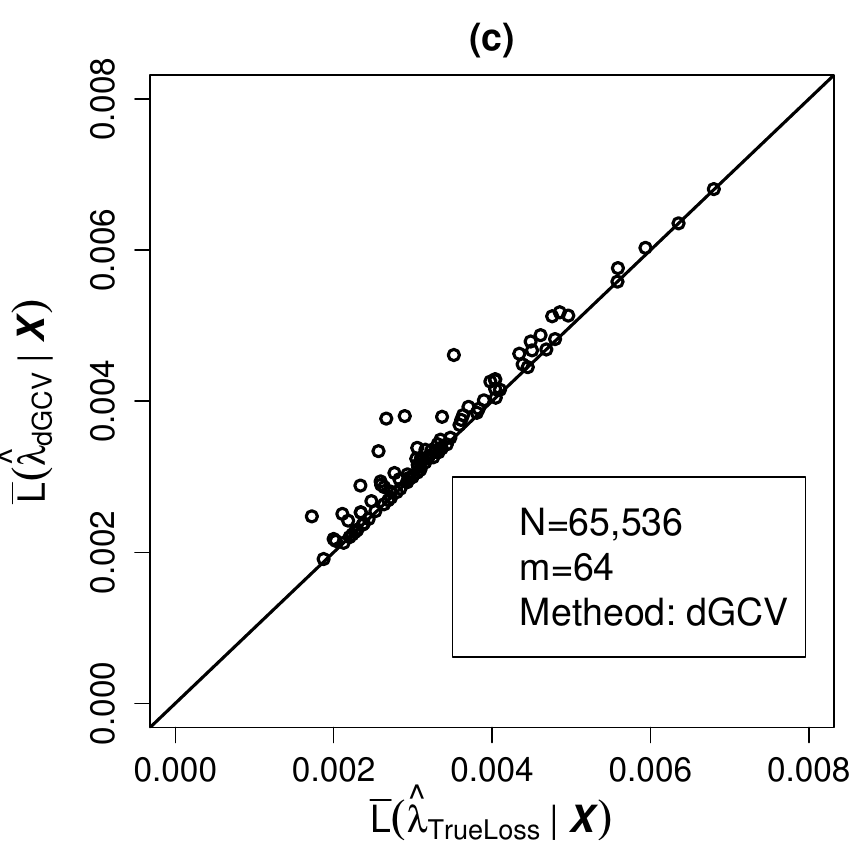}}
		\vspace{-2em}
	\end{center}
	\caption{(a) the logarithm of computational time (in seconds) v.s. $\log_2(m)$;  (b) the logarithm of averaged true empirical losses v.s. $\log_2(m)$;  (c) scatter plots of true empirical losses of function estimators.  Note that in (b), $\hat{\lambda}_{\rm opt}$ in the y-axis denotes one of $\hat{\lambda}_{\rm dGCV}$, $\hat{\lambda}_{\rm nGCV}$ and $\hat{\lambda}_{ \rm TrueLoss}$ for each curve. }
	\label{fig-6}
\end{figure}

To investigate the first issue, we simulated data from model~\eqref{simmod} with a sample size $N=2^{16}=65,536$ and the d-KRR was carried out using $m=2^j$ for $j=5,\cdots,11$. Summary statistics based on $100$ simulation runs are summarized in Figure~\ref{fig-6}, where the message is consistent with findings presented in Section~\ref{small}: at a much smaller computational cost, the d-KRR with a $\lambda$ chosen by minimizing $\dGCV$ is as good as using the $\lambda$ that minimizes the true empirical loss~\eqref{loss}, provided that the $m$ is not too large. 

\subsubsection{The Impact of the Choice of $m^*$}
\begin{figure}[htb!]
	\begin{center}
		\subfigure{\includegraphics[angle=0,width=0.5\textwidth,totalheight=0.5\textwidth]{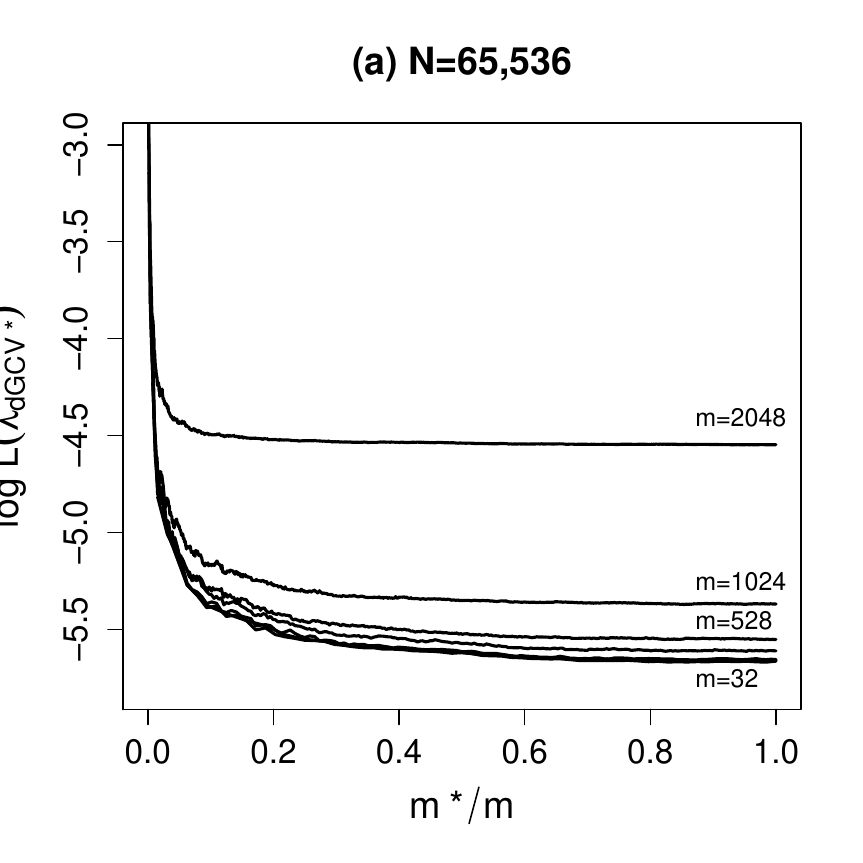}}
		\vspace{-2em}
	\end{center}
	\caption{ The logarithm of averaged true empirical losses v.s. $m^*/m$}
	\label{fig-7}
\end{figure}

When the sample size $N$ is large or even massive, it is inevitable to use a relative large $m$, in which case further computational savings can be achieved by choosing a subset of data for validation as suggested in~\eqref{gcv-m1} of Section~\ref{sec:cc}. The question remains that how small $m^*$ can be so that Theorem~\ref{thm1} still holds? As we have discussed in Remark~\ref{rem1}, a general rule of thumb for the choice of $m^*$ is that it cannot be too small compared to $m$. To shed some more lights on this issue, for each $m$, we simulate data from model~\eqref{simmod} and then fitted the d-KRR with the $\lambda$ that minimizes~\eqref{gcv-m1} using $m^*=1,\cdots,m$. Averaged empirical losses based on $100$ simulation runs are plotted in Figure~\ref{fig-7}, where it indicates that if $m^*$ is too small relative to $m$, the estimation accuracies indeed deteriorate significantly compared to the optimal performance. However, as long as $m^*$ is greater than $0.2 m$, the choice of $m^*$ has little impact on the estimation accuracies. Therefore, by setting $m^*$ as a reasonable percentage of $m$ (such as $20\%$ or $30\%$), one may indeed achieve a large reduction in computational cost without sacrificing too much on estimation accuracies. We want to emphasize again that it is worth to use a $m^*<m$ only when $N>>m^2$. And if used, whenever the computational cost is affordable, a larger $m^*$ is a safer choice to achieve better performances.

\section{The Million Song Dataset}\label{sec:rea}
In this section, we applied the $\dGCV^*$ tuning method to the Million Song Dataset, which consists of $463,715$ training examples and $51,630$ testing examples. Each observation is a song track released between the year 1922 and 2011. The response variable $y_i$ is the year when the song is released and the covariate $x_i$ is a $90$-dimensional vector, consists of timbre information of the song. We refer to \cite{BM11} for more details on this data set. Timbre is the quality of a musical note or sound that distinguishes different types of musical instruments, or voices \citep{jehan2011analyzer}. The goal is to use the timbre information of the song to predict the year when the song was released using the KRR. The same dataset has been analyzed by \cite{ZDW15}, but without addressing the issue of selecting an optimal tuning parameter. Our $\dGCV^*$ method demonstrated significant empirical advantages over theirs. 

\begin{figure}[htb!]
	\begin{center}
		\subfigure{\includegraphics[angle=0,width=0.5\textwidth,totalheight=0.5\textwidth]{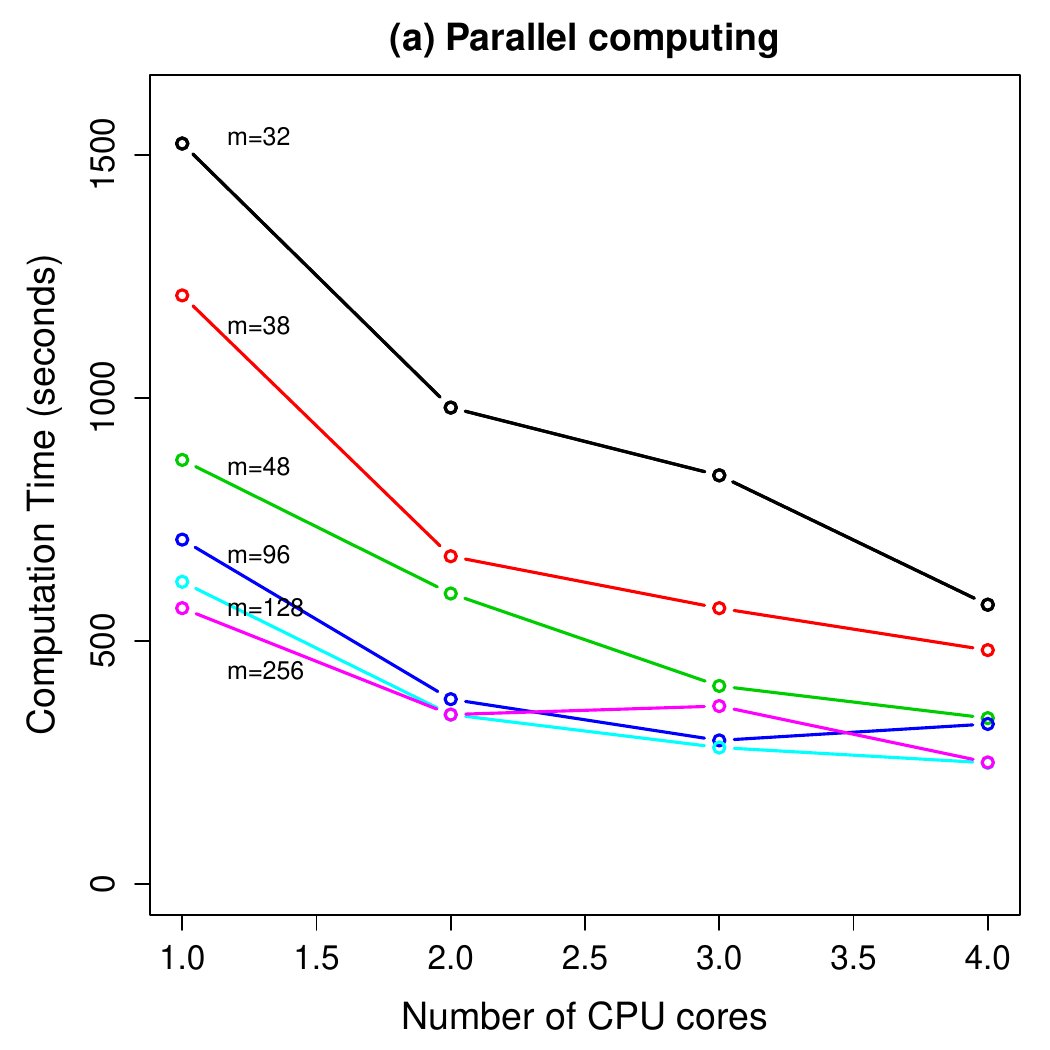}}
		\vspace{-2em}
	\end{center}
	\caption{ Computing times v.s. Number of CPU cores}
	\label{fig-7b}
\end{figure}

Following \cite{ZDW15}, the feature vectors were normalized so that they have mean $0$ and standard deviation $1$ and the Gaussian  kernel function $K(x,z)=\exp(-\|x-z\|_2^2/\phi)$ was used for the KRR. Seven partitions
$m\in \{32, 38, 48, 64, 96, 128, 256\}$ were used for the d-KRR.  Aside from the penalty parameter $\lambda$ in~(\ref{pls1}), the bandwidth $\phi$ is also known to have important impact on the prediction accuracy. To find the best combination of $(\lambda,\phi)$ for each partition $m$, we perform a $2$-dimensional search with $\lambda\in\{0.25,0.5,0.75,1.0,1.25,1.5\}/N$ and $\phi\in\{2,3,4,5,6,7\}$ by minimizing (\ref{gcv-m1}) with $m^*=\lceil m/10\rceil$, where $\lceil a\rceil$ is the smallest integer that is greater than $a$.  See Remark~\ref{rem3} for more details on the choice of $m^*$.  Note that in this case, $\dGCV^*(\lambda|\Xv)$ is also a function of $\phi$. The experiment was conducted in Matlab using a Windows desktop computer with 32GB of memory and a 2.6Ghz CPU with 4 CPU cores. To illustrate that the computation of the proposed  $\dGCV^*(\lambda|\Xv)$ can be easily paralleled, Figure~\ref{fig-7b} gives how averaged computation time changes as the number of CPU cores (in a single machine) increases. The computation time reduces most when the number of CPU cores increases from 1 to 2, and the reductions in computation times slow down as the number of CPU cores continues to increase. Such a trend is probably due to the memory constraints, communication costs and energy consumption limits on the computer and is not uncommon for parallel computing conducted in a single machine. Nevertheless, these computation times are reasonable for a data set with almost half-million observations and can be further reduced if a computing cluster is available.

\begin{figure}[htb!]
	\begin{center}
		\subfigure{\includegraphics[angle=0,width=0.32\textwidth,totalheight=0.3\textwidth]{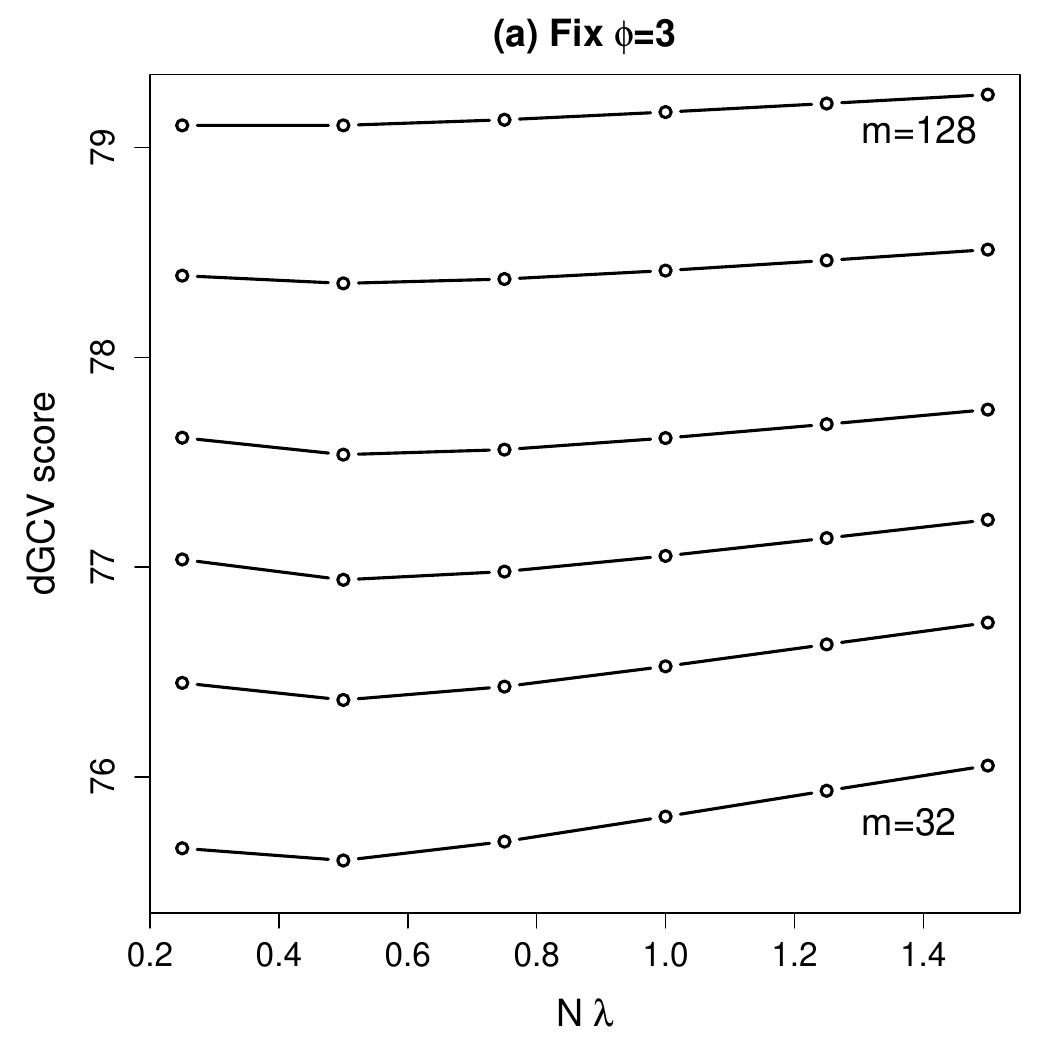}}
		\subfigure{\includegraphics[angle=0,width=0.32\textwidth,totalheight=0.3\textwidth]{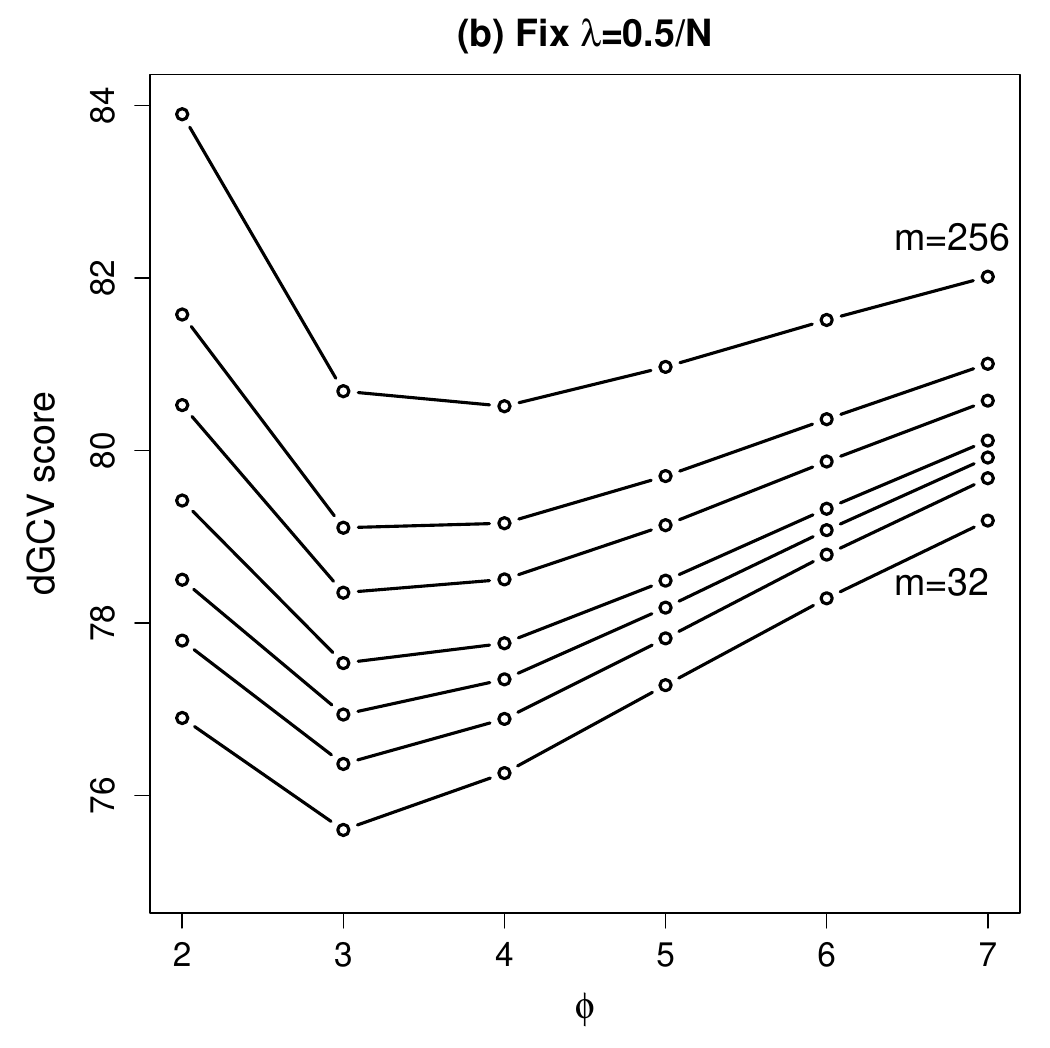}}
		\subfigure{\includegraphics[angle=0,width=0.32\textwidth,totalheight=0.3\textwidth]{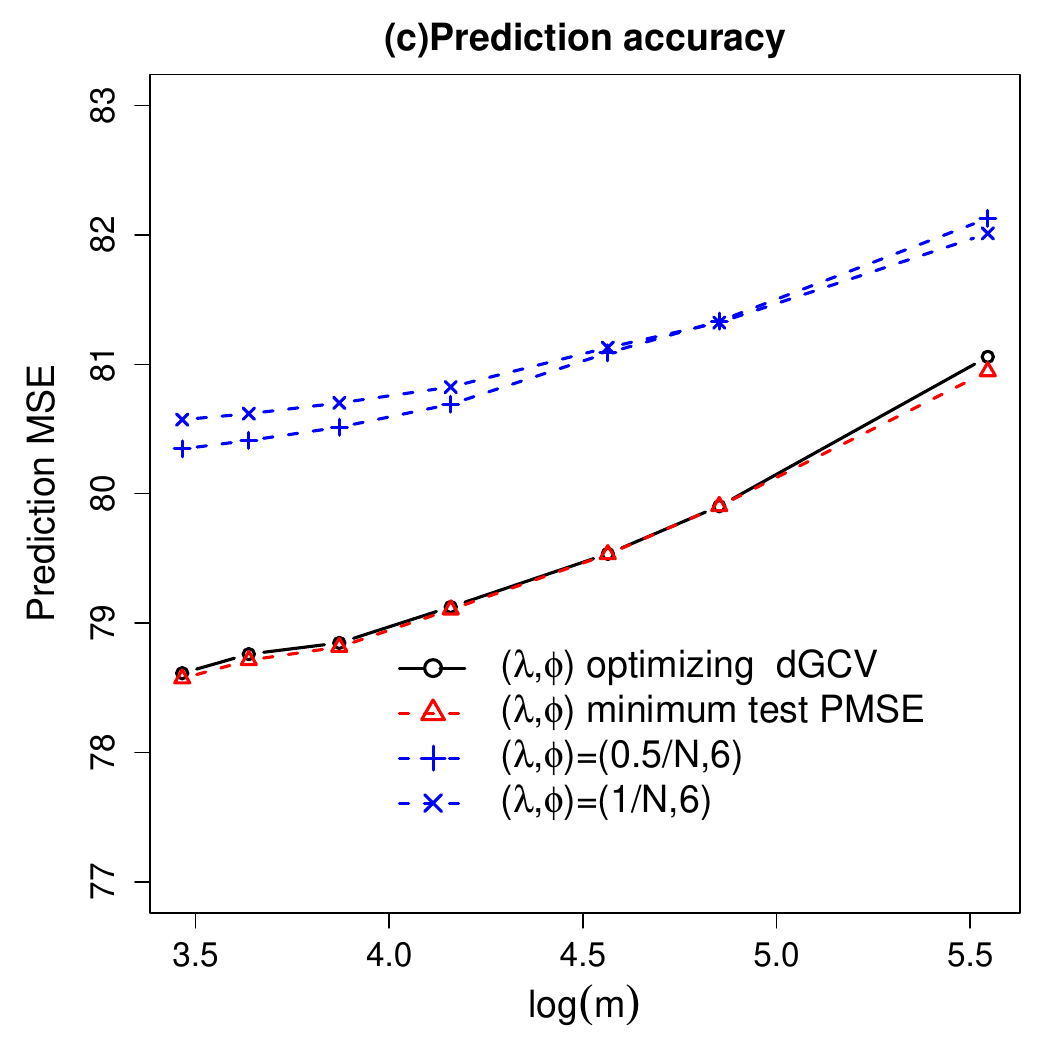}}
		\vspace{-2em}
	\end{center}
	\caption{(a) dGCV score v.s. $N\lambda$ with $m=32$ (the bottommost) to $m=128$ (the uppermost); (b) dGCV score v.s. $\phi$ with $m=32$ (the bottommost) to $m=256$ (the uppermost); (c) The prediction mean squared errors on the testing samples v.s. $\log(m)$. }
	\label{fig-2}
\end{figure}

The grid search gave the optimal choice of $\lambda=0.5/N$ and $\phi=3$ for most of case scenarios. From Figure~\ref{fig-2}(a)-(b), we can see that the choice of the bandwidth parameter $\phi$ has a great impacts on the dGCV$^*$ score as well as the penalty parameter $\lambda$. It seems that the latter provides some additional small adjustments after a good value of $\phi$ is chosen.

In~\cite{ZDW15}, the authors used a fixed value $\lambda=1/N$ and a $\phi=6$ chosen by the cross-validation for their kernel ridge regression model. In Figure~\ref{fig-2}(c), we can see that such a choice leads to a much worse prediction mean squared error (PMSE) on the testing samples. Using the proposed dGCV criterion, our choice of $\lambda$ and $\phi$ yields almost identical prediction accuracy as the minimum possible PMSE on the testing samples obtained over all $36$ grid points.

\begin{remark}
	\label{rem3}
	Note that for any given combination of $(\lambda,\phi)$, the estimated function $\bar{f}_{\lambda,\phi}$  used in ${\rm dGCV}^*$  is the same for different values of $m^*$, which is defined in~\eqref{ave}.
The agreement between the test PMSE of the ${\rm dGCV}^*$ method and the minimum test PMSE in Figure~\ref{fig-2}(c) suggests that there is no room to improve over the predictive performance of $\bar{f}_{\lambda,\phi}$ using tuning parameters selected by ${\rm dGCV}^*$, as long as the same multivariate Gaussian reproducing kernel function is used. This is a strong indication that $m^*=\lceil m/10\rceil$ is a good choice for this example, considering that ${\rm dGCV}^*$ did not use any information of the $51,630$ testing examples.
\end{remark}
\section{Discussion}
{
	In this paper, we proposed a data-driven criterion named dGCV that can be used to empirically selecting the critical tuning parameter $\lambda$ for d-KRR. Not only the proposed approach is computationally scalable even for massive data sets, we have also theoretically shown that it is asymptotically optimal in the sense that minimizing dGCV is equivalent to minimizing the true global conditional empirical loss, extending the existing optimality results of GCV to the divide-and-conquer framework.
	
	There are a few ways to extend the current work. For example, we have so far presumed a fixed $m$. One important direction is to investigate the growth rate of $m$ for some specific kernels under which Theorem~\ref{thm1} still holds, following the framework proposed in~\citet{SC17}. It is also of practical interest to develop a justifiable data-driven approach to detect the breaking point for $m$. Another interesting research direction is to develop a tuning criterion similar to the {\dGCV} for more general panelized Kernel regression such as \cite{zhang2016quantile} and \cite{chen2017double}. The definition of {\dGCV} in~\eqref{gcv-m} relies heavily on the closed form solution to the Kernel ridge regression, which is not available if the loss function or the penalty in~\eqref{pls1} are replaced by the quantile loss or the lasso penalty, respectively. The major difficulty lies in how to replace the effective degrees of freedom $\tr\left\{\bA_{kk}(\lambda)\right\}$'s in the denominator of~\eqref{gcv-m} when the hat matrices $\bA_{kk}$'s do not exist. Although there has been some research on this issue such as \cite{yuan2006gacv}, much more thorough investigations are needed.
}
\subsubsection*{Acknowledgements}
Ganggang Xu's research is partially supported by Collaboration Grants for Mathematicians (Award ID: 524205) from Simons Foundation and NSF SES-1758605. 
Zuofeng Shang's research is supported by NSF DMS-1764280.
Guang Cheng's research is partially supported by NSF CAREER Award DMS-1151692, DMS-1418042, DMS-1712907 and Office of Naval Research (ONR N00014-15-1-2331). 

\bibliographystyle{asa}
\bibliography{reference}

	\renewcommand{\thesection}{A}
	\section*{Appendix}

\pagestyle{empty}
\setcounter{equation}{0}
\setcounter{page}{1}
\renewcommand{\theequation}{A.\arabic{equation}}
\newtheorem{theo}{Theorem}[section]
\newtheorem{lem}[theo]{Lemma}
\setcounter{equation}{0}

From now on, we suppress the dependence of $\bA_{kl}(\lambda)$'s and $\bar{\bA}(\lambda)$ on $\lambda$ for ease of presentation and simply use $\bA_{kl}$'s and $\bar{\bA}$ whenever there is no ambiguity.
\begin{lem}\label{lemma:3:tec}
	\label{lemA1} Under the condition C1, we have that $\lambda_{max}(\bar{\bA}_{m}\bar{\bA}_{m}^T)=O_{\Px}(1)$.
\end{lem}
\begin{proof}
	Define the following matrix
	\[
	\begin{split}
	\bar{\bK}_m&=\frac{1}{m}\left(
	\begin{array}{cccc}
	\bK_{11} & \bK_{12}  & \cdots &\bK_{1m} \\
	\bK_{21} & \bK_{22}  & \cdots &\bK_{2m} \\
	\vdots & \vdots  & \ddots & \vdots \\
	\bK_{m1} & \bK_{m2}  & \cdots &\bK_{mm} \\
	\end{array}
	\right).
	\end{split}
	\]	
	Then it is straightforward to see that
	\[
	\bar{\bA}_{m}\bar{\bA}_{m}^T=\bar{\bK}\bD_1\bar{\bK}^T,
	\]
	where $\bD_1=\textrm{diag}\{\bB_{11},\dots,\bB_{mm}\}$ with $\bB_{ll}=(\bK_{ll}+n_l\lambda\bI_l)^{-2}$, for $l=1,\dots,m$. Then
	\[
	\bar{\bK}\bD_1\bar{\bK}^T=\frac{1}{m^2}\left(
	\begin{array}{c}
	\bK_{11} \\
	\bK_{21}\\
	\vdots  \\
	\bK_{m1}\\
	\end{array}
	\right)\bB_{11}(\bK_{11}^T,\dots,\bK_{m1}^T)+\cdots+\frac{1}{m^2}\left(
	\begin{array}{c}
	\bK_{1m} \\
	\bK_{2m}\\
	\vdots  \\
	\bK_{mm}\\
	\end{array}
	\right)\bB_{mm}(\bK_{1m}^T,\dots,\bK_{mm}^T),
	\]
	which implies that
	\[
	\begin{split}
	\lambda_{\max}(\bar{\bA}_{m}\bar{\bA}_{m}^T)&\leq\frac{1}{m^2}\sum_{l=1}^m\lambda_{\max}\{\left(
	\begin{array}{c}
	\bK_{1l} \\
	\bK_{2l}\\
	\vdots  \\
	\bK_{ml}\\
	\end{array}
	\right)\bB_{ll}(\bK_{1l}^T,\dots,\bK_{ml}^T)\}=\frac{1}{m^2}\sum_{l=1}^m\lambda_{\max}(\bB_{ll}\sum_{k=1}^m\bK_{kl}^T\bK_{kl})\\
	&=\frac{1}{m}\sum_{l=1}^m\lambda_{\max}\left\{(\bK_{ll}+n_l\lambda\bI_l)^{-2}\left(\frac{1}{m}\sum_{k=1}^m\bK_{kl}^T\bK_{kl}\right)\right\}=O_{\Px}(1).
	\end{split}
	\]
	The last inequality follows from condition C1. 
\end{proof}

\begin{lem}\label{lemma:4:tec}
	\label{lemA2} Under the conditions C1-C2 and C3(a), for a fixed $\lambda$, we have that
	\be
	\label{eq:lemA2}
	\bar{L}(\lambda|\Xv)-\bar{R}(\lambda|\Xv)=o_{\Pepsx}\{\bar{R}(\lambda|\Xv)\}.
	\ee
\end{lem}
\begin{proof}
	Using similar notations in equation~(\ref{risk2}), it is straightforward to show that
	\be
	\label{loss2}
	\bar{L}(\lambda|\Xv)=\frac{1}{N}\left(\bar{\bA}_m\Yv-\Fv\right)^T\bW\left(\bar{\bA}_m\Yv-\Fv\right), \text{ with } \Yv=\Fv+\beps.
	\ee
	Using (\ref{risk2}), we have that
	\[
	\bar{L}(\lambda|\Xv)-\bar{R}(\lambda|\Xv)=-\frac{2}{N}\Fv^T(\bI-\bar{\bA}_m)^T\bW\bar{\bA}_m\beps+\frac{1}{N}\beps^T\bar{\bA}_m^T\bW\bar{\bA}_m\beps-\frac{\sigma^2}{N}\tr(\bar{\bA}_m^T\bW\bar{\bA}_m).
	\]
	Since the random error $\varepsilon$ and the covariate $X$ are independent in model~(\ref{model}), to show~(\ref{eq:lemA2}), it suffices to show the following two equations
	\begin{eqnarray}
	\Vare\left\{\frac{1}{N}\Fv^T(\bI-\bar{\bA}_m)^T\bW\bar{\bA}_m\beps\right\}=o_{\Px}\{\bar{R}^2(\lambda|\Xv)\},\label{eq:lemA2:1}\\
	\Vare\left\{\frac{1}{N}\beps^T\bar{\bA}_m^T\bW\bar{\bA}_m\beps-\frac{\sigma^2}{N}\tr(\bar{\bA}_m^T\bW\bar{\bA}_m)\right\}=o_{\Px}\{\bar{R}^2(\lambda|\Xv)\}. \label{eq:lemA2:2}
	\end{eqnarray}
	We first show (\ref{eq:lemA2:1}). Straightforward algebra yields that
	\[
	\begin{split}
	\Vare\left\{\frac{1}{N}\Fv^T(\bI-\bar{\bA}_m)^T\bW\bar{\bA}_m\beps\right\}&=\frac{\sigma^2}{N^2}\Fv^T(\bI-\bar{\bA}_m)^T\bW\left(\bar{\bA}_m\bar{\bA}_m^T\right)\bW(\bI-\bar{\bA}_m)\Fv\\
	&\leq \frac{\sigma^2\lambda_{\max}\left(\bar{\bA}_m\bar{\bA}_m^T\bW\right)}{N}\frac{1}{N}\Fv^T(\bI-\bar{\bA}_m)^T\bW(\bI-\bar{\bA}_m)\Fv\\
	&\leq \frac{\sigma^2\lambda_{\max}\left(\bar{\bA}_m\bar{\bA}_m^T\right)\lambda_{\max}(\bW)}{N\bar{R}(\lambda|\Xv)}\bar{R}^2(\lambda|\Xv)\\
	&=o_{\Px}(1)\bar{R}^2(\lambda|\Xv)=o_{\Px}\{\bar{R}^2(\lambda|\Xv)\},
	\end{split}
	\]
	where the second last equation follows from conditions C2 and C3(a) and Lemma~(\ref{lemA1}). 
	
	Now we show (\ref{eq:lemA2:2}).  Straightforward algebra yields that
	\be
	\label{A.4part}
	\begin{split}
		\Vare&\left\{\frac{1}{N}\beps^T\bar{\bA}_m^T\bW\bar{\bA}_m\beps-\frac{\sigma^2}{N}\tr(\bar{\bA}_m^T\bW\bar{\bA}_m)\right\}=\frac{\EE\varepsilon^4-\sigma^4}{N^2}\sum_{i=1}^{N}\bar{b}_{ii}^2+2\sigma^4\sum_i\sum_{j}^{i\neq j}b_{ij}^2\\
		&\leq\frac{K_1}{N^2}\tr\{(\bar{\bA}_m^T\bW\bar{\bA}_m)^2\}
		\leq\frac{K_1\lambda_{\max}(\bar{\bA}_m^T\bW\bar{\bA}_m)}{N^2}\tr(\bar{\bA}_m^T\bW\bar{\bA}_m)\\&\leq \frac{K_1\lambda_{\max}(\bar{\bA}_m^T\bW\bar{\bA}_m)}{N\sigma^2}\bar{R}(\lambda|\Xv)\leq \frac{K_1\lambda_{\max}(\bar{\bA}_m^T\bar{\bA}_m)\lambda_{\max}(\bW)}{\sigma^2N\bar{R}(\lambda|\Xv)}\bar{R}^2(\lambda|\Xv)\\&=o_{\Px}(1)\bar{R}^2(\lambda|\Xv)\\
	\end{split}
	\ee
	where $\bar{b}_{ij}$ is the $(i,j)$th element of matrix $\bar{\bA}_m^T\bW\bar{\bA}_m$ and $K_1=\EE\varepsilon^4+\sigma^4$. The last equality follows from conditions C2 and C3(a) and Lemma~\ref{lemA1}. Using~(\ref{eq:lemA2:1})-(\ref{eq:lemA2:2}), the equation~(\ref{eq:lemA2}) follows from a simple application of the Cauchy-Schwartz inequality and the Markov's inequality. The proof is complete.
\end{proof}

\begin{proof}[{\bf Proof of Lemma~\ref{lem1}}]
	Using~(\ref{loss2}) and (\ref{U}), we have that
	\be
	\label{eq:lem1:1}
	\begin{split}
		\bar{U}(\lambda|\Xv)-\bar{L}(\lambda|\Xv)-\frac{1}{N}\beps^T\bW\beps&=\frac{2}{N}\Fv^T(\bI-\bar{\bA}_m)^T\bW\beps-\frac{2}{N}\left\{\beps^T\bar{\bA}_m\bW\beps-\sigma^2\tr(\bar{\bA}_m\bW)\right\}.
	\end{split}
	\ee
	Notice that the random error $\varepsilon$ and the covariate $X$ are independent in model~(\ref{model}).
	We will show~(\ref{eq:lem1}) using equation~(\ref{eq:lemA2}) in Lemma~\ref{lemA2}, for which
	it suffices to show the following two equations
	\begin{eqnarray}
	\Vare\left\{\frac{1}{N}\Fv^T(\bI-\bar{\bA}_m)^T\bW\beps\right\}=o_{\Px}\{\bar{R}^2(\lambda|\Xv)\},\label{eq:lem1:2}\\
	\Vare\left\{\frac{1}{N}\beps^T\bar{\bA}_m\bW\beps-\frac{\sigma^2}{N}\tr(\bar{\bA}_m\bW)\right\}=o_{\Px}\{\bar{R}^2(\lambda|\Xv)\}.\label{eq:lem1:3}
	\end{eqnarray}
	We first show (\ref{eq:lem1:2}). Straightforward algebra yields that
	\[
	\begin{split}
	\Vare\left\{\frac{1}{N}\Fv^T(\bI-\bar{\bA}_m)^T\bW\beps\right\}&=\frac{\sigma^2}{N^2}\Fv^T(\bI-\bar{\bA}_m)^T\bW^2(\bI-\bar{\bA}_m)\Fv\leq \frac{\sigma^2\lambda_{\max}(\bW)}{N\bar{R}(\lambda|\Xv)}\bar{R}^2(\lambda|\Xv)\\
	&=o_{\Px}(1)\bar{R}^2(\lambda|\Xv)=o_{\Px}\{\bar{R}^2(\lambda|\Xv)\},
	\end{split}
	\]
	where the second last equation follows from conditions C2-C3. Next, we show (\ref{eq:lem1:3}). Using condition C2, similar to the inequality~\eqref{A.4part}, it is straightforward to show that
	\[
	\begin{split}
	\Vare\left\{\frac{1}{N}\beps^T\bar{\bA}_m\bW\beps\right\}&\leq \frac{K_1}{N^2}\tr(\bar{\bA}_m^T\bW^2\bar{\bA}_m)\leq \frac{K_1\lambda_{\max}(\bW)}{N\sigma^2}\bar{R}(\lambda|\Xv)\\
	&=\frac{K_1\lambda_{\max}(\bW)}{\sigma^2N\bar{R}(\lambda|\Xv)}\bar{R}^2(\lambda|\Xv)=o_{\Px}(1)\bar{R}^2(\lambda|\Xv),
	\end{split}
	\]
	where  $K_1=\EE\varepsilon^4+\sigma^4$ is bounded. Hence, (\ref{eq:lem1:3}) is proved using, again, condition C2-C3.  Using~(\ref{eq:lem1:2})-(\ref{eq:lem1:3}) and (\ref{eq:lemA2}), the equation~(\ref{eq:lem1}) follows from a simple application of the Cauchy-Schwartz inequality and the Markov's inequality. The proof is complete.
\end{proof}

\begin{proof}[{\bf Proof of Theorem~\ref{thm1} }] Using Lemma~\ref{lem1} and Lemma~\ref{lemA2}, it suffices to show that
	\be
	\label{difGU}
	\dGCV_{DC}(\lambda|\Xv)-\bar{U}(\lambda|\Xv)=o_{\Pepsx}\{\bar{R}(\lambda|\Xv)\}.
	\ee
	Using the first order Taylor expansion of $(1-x)^{-2}$ around $x=0$, we have that $(1-x)^{-2}=1+2x+3(1-x^*)^{-4}x^2$ for some $x^*\in (0,x)$. Under condition C3, we have that $\frac{\tr(\bar{\bA}_m)}{N}=o_{\Px}(1)$ and thus we can consider the following decomposition
	\[
	\begin{split}
	\dGCV(\lambda|\Xv)-\bar{U}(\lambda|\Xv)=&\underbrace{\left\{\frac{1}{N}\Yv^T\{\bI-\bar{\bA}_m(\lambda)\}^T\bW\{\bI-\bar{\bA}_m(\lambda)\}\Yv-\sigma^2\right\}\frac{2\tr(\bar{\bA}_m\bW)}{N}}_{I}\\&+\underbrace{\frac{1}{N}\Yv^T\{\bI-\bar{\bA}_m(\lambda)\}^T\bW\{\bI-\bar{\bA}_m(\lambda)\}\Yv O_{\Px}\left(\frac{\{\tr(\bar{\bA}_m\bW)\}^2}{N^2}\right)}_{II}
	\end{split}
	\]
	Using condition C4, we have that
	\be
	\label{trace}
	\frac{\tr(\bar{\bA}_m\bW)}{N}=o_{\Px}\{\bar{R}^{1/2}(\lambda|\Xv)\},
	\ee
	which implies that $II=o_{\Px}(\bar{R}(\lambda|\Xv))$ since $\frac{1}{N}\Yv^T\{\bI-\bar{\bA}_m(\lambda)\}^T\bW\{\bI-\bar{\bA}_m(\lambda)\}\Yv$ is bounded. For part $I$, we can write
	\[
	\begin{split}
	I&=\left\{\frac{1}{N}\Yv^T\{\bI-\bar{\bA}_m(\lambda)\}^T\bW\{\bI-\bar{\bA}_m(\lambda)\}\Yv-\sigma^2\right\}\frac{2\tr(\bar{\bA}_m\bW)}{N}\\&=\left\{\bar{U}(\lambda|\Xv)-\frac{1}{N}\beps^T\bW\beps\right\}\frac{2\tr(\bar{\bA}_m\bW)}{N}
	+\left(\frac{1}{N}\beps^T\bW\beps-\sigma^2\right)\frac{2\tr(\bar{\bA}_m\bW)}{N}-\frac{4\{\tr(\bar{\bA}_m\bW)\}^2\sigma^2}{N^2}.
	\end{split}
	\]
	By Lemma~\ref{lem1}, we have that $\bar{U}(\lambda|\Xv)-\frac{1}{N}\beps^T\bW\beps=\bar{R}(\lambda|\Xv)+o_{\Pepsx}\{\bar{R}(\lambda|\Xv)\}$. Under condition C3, one has that $\frac{\tr(\bar{\bA}_m\bW)}{N}=o_{\Px}(1)$, and thus
	\[
	\left\{\bar{U}(\lambda|\Xv)-\frac{1}{N}\beps^T\bW\beps\right\}\frac{2\tr(\bar{\bA}_m\bW)}{N}=o_{\Pepsx}\{\bar{R}(\lambda|\Xv)\}.
	\]
	Furthermore, since $\frac{1}{N}\beps^T\bW\beps-\sigma^2=O_{\Peps}(N^{-1/2})$ (condition C3 (a)) and  
	$N\bar{R}(\lambda|\Xv){\xrightarrow{\Px}}\infty$ (condition C2), we have that $\frac{1}{N}\beps^T\bW\beps-\sigma^2=o_{\Pepsx}\{\bar{R}^{1/2}(\lambda|\Xv)\}$. Using this and equation~(\ref{trace}), we have that
	\[
	\left(\frac{1}{N}\beps^T\bW\beps-\sigma^2\right)\frac{2\tr(\bar{\bA}_m\bW)}{N}=o_{\Pepsx}\{\bar{R}(\lambda|\Xv)\}.
	\]
	The third part of $I$ is $o_{\Px}\{\bar{R}(\lambda|\Xv)\}$ due to equation~(\ref{trace}). Therefore, we have shown that
	\[
	\dGCV(\lambda|\Xv)-\bar{U}(\lambda|\Xv)=o_{\Pepsx}\{\bar{R}(\lambda|\Xv)\},
	\]
	which completes the proof.
\end{proof}

\begin{lem}\label{prop:entropy} Define the following class of non-negative functions 
	\be
	\mathcal{F}=\{f\in L_2(\Pb):f\geq 0, \|f\|_{\sup}\leq V, J_1(f)\leq V^2H^2\}, 
	\ee
	where $V>0$ and $H>0$ are constants.
	If condition C4'(d) holds and
	$(\epsilon_n,\gamma_n)$ satisfy
	\begin{equation}\label{additional:condition}
	\epsilon_n^3\gamma_n^2\geq \frac{c_0(1+H)V}{n},
	\end{equation}
	where $c_0>0$ is a constant,
	then there exists a constant $C>0$ such that
	for all $n$,
	\begin{eqnarray*}
		P\left(\sup_{f\in\mathcal{F}}\frac{|\Pb_n f-\Pb f|}{\Pb_n f+\Pb f+\gamma_n(\Pb_n f+\Pb f+1)}>C\epsilon_n\right)\le \exp(-n\epsilon_n^2\gamma_n/2).
	\end{eqnarray*} 
\end{lem}

\begin{proof}
	Recall the definition of $\mathcal{F}_0$ in condition C4'(d). It can be checked that
	\[
	\mathcal{F}\subseteq V(1+H)\mathcal{F}_0.
	\]
	Hence under condition C4'(d), we have that with probability approaching one,
	\begin{eqnarray*}
		N(\epsilon_n\gamma_n,\|\cdot\|_{\Pb_n},\mathcal{F})&
		\le& N(\epsilon_n\gamma_n,\|\cdot\|_{\Pb_n},V(1+H)\mathcal{F}_0)
		=N\left(\frac{\epsilon_n\gamma_n}{V(1+H)},\|\cdot\|_{\Pb_n},\mathcal{F}_0\right)\\&\leq&\exp\left\{\frac{C_0(1+H)V}{\epsilon_n\gamma_n}\right\}.
	\end{eqnarray*}
	
	By the Theorem given in \cite{P95} and the Theorem~2.1 of \cite{P86}, there exists constants $C$ and $c_0$ such that	
	\begin{eqnarray*}
		P\left(\sup_{f\in\mathcal{F}}\frac{|\Pb_n f-\Pb f|}{\Pb_n f+\Pb f+\gamma_n(\Pb_n f+\Pb f+1)}>C\epsilon_n\right)
		&\le& \exp\left(c_0\frac{(1+H)V}{2\epsilon_n\gamma_n}-n\epsilon_n^2\gamma_n\right)\\
		&\le&\exp(-n\epsilon_n^2\gamma_n/2). 
	\end{eqnarray*}
\end{proof}

\begin{proof}[{\bf Proof of Lemma~\ref{lem3}}]
	Define the kernel matrix	\[
	\begin{split}
	\bK&=\left(
	\begin{array}{cccc}
	\bK_{11} & \bK_{12}  & \cdots &\bK_{1m} \\
	\bK_{21} & \bK_{22}  & \cdots &\bK_{2m} \\
	\vdots & \vdots  & \ddots & \vdots \\
	\bK_{m1} & \bK_{m2}  & \cdots &\bK_{mm} \\
	\end{array}
	\right)=\bPhi\bPhi^T,
	\end{split}
	\]	
	where $\bPhi$ is a $N\times r$ matrix with $r$ being the rank of $\bK$. By this notation, we have that
	\[
	\bK_{ll}=\bPhi_l\bPhi_l^T,\qquad l=1,\cdots,m.
	\]
	where $\bPhi_l$ is a $n_l\times r$ submatrix of $\bPhi$ consists of rows corresponding to a subdata set $S_l$.  Then it is straightforward to show that
	\[
	\lambda_{\max}\left\{(\bK_{ll}+n_l\lambda\bI_l)^{-2}\left(\frac{1}{m}\sum_{k=1}^m\bK_{kl}^T\bK_{kl}\right)\right\}=\frac{1}{m}\lambda_{\max}\left\{\bPhi\bPhi_l^T(\bPhi_l\bPhi_l^T+n_l\lambda\bI_l)^{-2}\bPhi_l\bPhi^T\right\}.
	\]
	Using the  Sherman–Morrison formula, we can show that
	\[
	\begin{split}
	\bPhi_l^T(\bPhi_l\bPhi_l^T+n_l\lambda\bI_l)^{-1}&=\bPhi_l^T\left[n^{-1}\lambda^{-1}\bI-n^{-2}\lambda^{-2}\bPhi_l(\bI+n^{-1}\lambda^{-1}\bPhi_l^T\bPhi_l)^{-1}\bPhi_l^T\right]\\&=n^{-1}\lambda^{-1}\left[\bI-n^{-1}\lambda^{-1}\bPhi_l^T\bPhi_l(\bI+n^{-1}\lambda^{-1}\bPhi_l^T\bPhi_l)^{-1}\right]\bPhi_l^T\\
	&=n^{-1}\lambda^{-1}\left[(\bI+n^{-1}\lambda^{-1}\bPhi_l^T\bPhi_l)^{-1}\right]\bPhi_l^T,
	\end{split}
	\]
	which gives that
	\[
	\begin{split}
	&\lambda_{\max}\left\{(\bK_{ll}+n_l\lambda\bI_l)^{-2}\left(\frac{1}{m}\sum_{k=1}^m\bK_{kl}^T\bK_{kl}\right)\right\}\\&=\frac{n^{-2}\lambda^{-2}}{m}\lambda_{\max}\left\{\bPhi\left[(\bI+n^{-1}\lambda^{-1}\bPhi_l^T\bPhi_l)^{-1}\right]\bPhi_l^T\bPhi_l\left[(\bI+n^{-1}\lambda^{-1}\bPhi_l^T\bPhi_l)^{-1}\right]\bPhi^T\right\}
	\\&=\frac{n^{-1}\lambda^{-1}}{m}\lambda_{\max}\left\{\bPhi(\bI+n^{-1}\lambda^{-1}\bPhi_l^T\bPhi_l)^{-1}\bPhi^T-\bPhi(\bI+n^{-1}\lambda^{-1}\bPhi_l^T\bPhi_l)^{-2}\bPhi^T\right\}\\
	&\leq \frac{1}{N}\lambda_{\max}\left\{\bPhi(\lambda\bI+n^{-1}\bPhi_l^T\bPhi_l)^{-1}\bPhi^T\right\}= \lambda_{\max}\left\{(\lambda\bI+n^{-1}\bPhi_l^T\bPhi_l)^{-1}\left[\frac{1}{N}\bPhi^T\bPhi\right]\right\}.\\
	\end{split}
	\]
	Using the following identity from the Appendix B of \cite{Bach13}
	\[
	\begin{split}
	(\lambda\bI+n^{-1}\bPhi_l^T\bPhi_l)^{-1}&=\left(\lambda\bI+\frac{1}{N}\bPhi^T\bPhi-\frac{1}{N}\bPhi^T\bPhi+n^{-1}\bPhi_l^T\bPhi_l\right)^{-1}\\&=\left(\lambda\bI+\frac{1}{N}\bPhi^T\bPhi\right)^{-1/2}\left[\bI-\frac{1}{N}\bPsi^T\bPsi+\frac{1}{n}\bPsi_l^T\bPsi_l\right]^{-1}\left(\lambda\bI+\frac{1}{N}\bPhi^T\bPhi\right)^{-1/2},
	\end{split}
	\]
	where $\bPsi=\bPhi\left(\lambda\bI+\frac{1}{N}\bPhi^T\bPhi\right)^{-1/2}$ and $\bPsi_l$ is the submatrix of $\bPsi$, we have that 
	\[
	\begin{split}
	&\lambda_{\max}\left\{(\bK_{ll}+n_l\lambda\bI_l)^{-2}\left(\frac{1}{m}\sum_{k=1}^m\bK_{kl}^T\bK_{kl}\right)\right\}\leq \lambda_{\max}\left\{(\lambda\bI+n^{-1}\bPhi_l^T\bPhi_l)^{-1}\left[\frac{1}{N}\bPhi^T\bPhi\right]\right\}\\
	&=\lambda_{\max}\left\{\left(\lambda\bI+\frac{1}{N}\bPhi^T\bPhi\right)^{-1/2}\left[\bI-\frac{1}{N}\bPsi^T\bPsi+\frac{1}{n}\bPsi_l^T\bPsi_l\right]^{-1}\left(\lambda\bI+\frac{1}{N}\bPhi^T\bPhi\right)^{-1/2}\left[\frac{1}{N}\bPhi^T\bPhi\right]\right\}\\
	&\leq \sigma_{\max}\left\{\left[\bI-\frac{1}{N}\bPsi^T\bPsi+\frac{1}{n}\bPsi_l^T\bPsi_l\right]^{-1}\right\}\lambda_{\max}\left\{\left(\lambda\bI+\frac{1}{N}\bPhi^T\bPhi\right)^{-1/2}\left[\frac{1}{N}\bPhi^T\bPhi\right]\left(\lambda\bI+\frac{1}{N}\bPhi^T\bPhi\right)^{-1/2}\right\}\\
	&\leq \sigma_{\max}\left\{\left[\bI-\frac{1}{N}\bPsi^T\bPsi+\frac{1}{n}\bPsi_l^T\bPsi_l\right]^{-1}\right\},
	\end{split}
	\]
		where $\sigma_{\max}(\bA)$ is the spectral norm of the matrix $\bA$.
		
	Therefore, to show condition C1, it suffices to show that
	\be
	\label{key1}
	\max_{l=1,\cdots,m}\lambda_{\max}\left[\frac{1}{N}\bPsi^T\bPsi-\frac{1}{n}\bPsi_l^T\bPsi_l\right]=o_{\Px}(1).
	\ee
	
	Using Lemma~2 of ~\cite{Bach13}, we have that
	\be
	\label{ineq}
	\mathbb P_{I}\left(\lambda_{\max}\left[\frac{1}{N}\bPsi^T\bPsi-\frac{1}{n}\bPsi_l^T\bPsi_l\right]>t\right)\leq r\exp\left(\frac{-nt^2/2}{\lambda_{\max}\left[\frac{1}{N}\bPsi^T\bPsi\right](R^2+t/3)}\right),
	\ee
	where $\mathbb P_{I}$ is the probability measure corresponding to the partition of the data, $r={\rm rank}(\bPsi)={\rm rank}(\bK)$ and $R$ is the upperbound of L2-norm of all rows of $\bPsi$. In our case, L2-norm of all rows of $\bPsi$ the diagonal elements of matrix
	\[
	\bPsi\bPsi^T=\bPhi\left(\lambda\bI+\frac{1}{N}\bPhi^T\bPhi\right)^{-1}\bPhi^T=N\bK(\bK+N\lambda\bI)^{-1},
	\]
	where the last equality follows from the  Sherman–Morrison formula. Then, by the definition of $d_{\lambda}$ in~\eqref{mdf}, we have that $R^2\leq d_{\lambda}$. In addition, note that
	\[
	\lambda_{\max}\left(\frac{1}{N}\bPsi^T\bPsi\right)=\lambda_{\max}\left(\frac{1}{N}\bPhi\left(\lambda\bI+\frac{1}{N}\bPhi^T\bPhi\right)^{-1}\bPhi^T\right)\leq 1,
	\]
	which implies that inequality~\eqref{ineq} can be further simplified as
	\[
	\mathbb P_{I}\left(\lambda_{\max}\left[\frac{1}{N}\bPsi^T\bPsi-\frac{1}{n}\bPsi_l^T\bPsi_l\right]>t\right)\leq r\exp\left(\frac{-nt^2/2}{d_{\lambda}+t/3}\right),
	\]
	which further leads to that 
		\[
		\begin{split}
\mathbb P_{I}\left(\max_{l=1,\cdots,m}\lambda_{\max}\left[\frac{1}{N}\bPsi^T\bPsi-\frac{1}{n}\bPsi_l^T\bPsi_l\right]>t\right)
\leq mr\exp\left(\frac{-nt^2/2}{d_{\lambda}+t/3}\right)\rightarrow^{\Px} 0,
		\end{split}
		 \]
	
	for any $0<t<3d_{\lambda}$  under condition C1', which completes the proof of~\eqref{key1}.
\end{proof}
\begin{proof}[{\bf Proof of Lemma~\ref{lem2}}]
	We first consider $Q_2(\lambda|\Xv)$ in~(\ref{Q2}). Define the function class 
	\[
	\mathcal{F}_n=\left\{f(x): \|f\|_{\sup}\leq C_1V_n,J_1(f)\leq C_2V_n^2H_n^2\right\},
	\] 
	where $V_n$ and $H_n$ are as defined in Conditions C4'(b)-(c)  and $C_1,C_2$ are some constants. Applying Lemma~\ref{prop:entropy} to the function class $\mathcal{F}_n$ with $\epsilon_n=\epsilon$ and $\gamma_n=\sqrt{\frac{c_0(1+H_n)V_n}{n}}$, which satisfy (\ref{additional:condition}) under Conditions C4'(b)-(c), we have that
	\be
	\label{eq:emp1}
	P\left(\sup_{f\in\mathcal{V}_n}\frac{|\Pb_n f-\Pb  f|}{\Pb_n  f+\Pb  f+\gamma_n}>C\epsilon\right)\leq \exp(-n\epsilon^2\gamma_n/2). 
	\ee

	Let 
	$v_k(x)=\Vare\left\{\widehat{f}_k(x)\right\}$, $k=1,\dots,m$. It is straightforward to show that the first derivative of $v_k(x)$ are bounded as follows
	\[
	|v'_k(x)|=2\left|\Cove\left\{\widehat{f}_k(x), \widehat{f}'_k(x)\right\}\right|\leq 2\sqrt{v_k(x)}\sqrt{\Vare\{\widehat{f}'_k(x)\}},
	\]
	which further implies that
	\[
	\begin{split}
	J_1(v_k)&=\int_{\mathcal{X}}\{v'_k(x)\}^2\,d\Px(x)\leq 4\|v_k\|_{\sup}\int_{\mathcal{X}}\Vare\{\widehat{f}'_k(x)\}\,d\Px(x)\\
	&\leq 4\|v_k\|_{\sup}^2\frac{\int_{\mathcal{X}}\Vare\{\widehat{f}'_k(x)\}\,d\Px(x)}{\int_{\mathcal{X}}v_k(x)\,d\Px(x)}=O_{\Px} (V_n^2H_n^2).
	\end{split}
	\]		
	Therefore, under conditions C4'(a)-(b), we have that
	\[
	v_1(x),\dots,v_m(x)\in \mathcal{F}_n\text{ in probability measure $\Px$.}
	\]
	For simplicity, from now on, we use $Q$ for $Q(\lambda|\Xv)$ in~(\ref{Q}) and $Q_j$ for $Q_j(\lambda|\Xv)$, $j=1,2$, in~(\ref{Q1}) and (\ref{Q2}) whenever there is no ambiguity.
	Using the facts that $Q=\frac{1}{m^2}\sum_{k=1}^{m}\Pb v_k$ and $Q_2=\frac{1}{m^2}\sum_{k=1}^{m}\Pb_{n_k}v_k$, a direct application of~(\ref{eq:emp1}) gives that
	\[
	\begin{split}
	P\left(\frac{|Q_2-Q|}{Q_2+Q+\frac{1}{m}\gamma_n}>C\epsilon\right)&\leq P\left(\frac{\frac{1}{m}\sum_{k=1}^{m}|\Pb_{n_k}v_k-\Pb_{n_k}v_k|}{\frac{1}{m}\sum_{k=1}^{m}(\Pb_{n_k}v_k+\Pb_{n_k}v_k)+\gamma_n}>C\epsilon\right)\\
	&\leq P\left(\max_{1\le k\le m}\left(\frac{|\Pb_{n_k}v_k-\Pb_{n_k}v_k|}{\Pb_{n_k}v_k+\Pb_{n_k}v_k+\gamma_n}\right)>C\epsilon\right)\\
	&  \leq m\exp(-n\epsilon^2\gamma_n/2)\rightarrow 0,
	\end{split}
	\]
	where the last step follows from condition C4'(c). In addition, by conditions C4'(b)-(c), we have that $\frac{\gamma_n}{mQ}=\sqrt{\frac{c_0(1+H_n)V_n}{mNQ^2}}=O_{\Px}(1)$. Hence we conclude that
	\be
	\label{eq:Q2}
	Q_2(\lambda|\Xv)=Q(\lambda|\Xv)+o_{\Px}Q\{(\lambda|\Xv)\}.
	\ee
	Now we turn to the quantity $Q_1(\lambda|\Xv)$. Define another function class
	\[
	\bar{\mathcal{F}}_n=\left\{f(x): \|f\|_{\sup}\leq C_1\frac{V_n}{m},J_1(f)\leq C_2\frac{V_n^2H_n^2}{m^2}\right\},
	\] 
	where $V_n$ and $H_n$ are as defined in Conditions C4'(b)-(c)  and $C_1,C_2$ are some constants.  By applying Lemma~\ref{prop:entropy} to the function class $\bar{\mathcal{F}}_n$ with $\epsilon_n=\epsilon$ and $\gamma_N=\sqrt{\frac{c_0(1+H_n)V_n}{mN}}$, which satisfy (\ref{additional:condition}) under Conditions C4'(b)-(c), we have that
	\be
	\label{eq:emp2}
	P\left(\sup_{f\in\mathcal{V}_N}\frac{|\Pb_N f-\Pb  f|}{\Pb_N  f+\Pb  f+\gamma_N}>C\epsilon\right)\leq \exp(-N\epsilon^2\gamma_N/2). 
	\ee
	Define another function
	\[
	\bar{v}(x)=\Vare\{\bar{f}(x)\}=\frac{1}{m^2}\sum_{k=1}^{m}v_k(x),
	\]
	whose derivative is bounded as
	\[
	\begin{split}
	|\bar{v}'(x)|&=2\left|\Cove\left\{\bar{f}(x), \bar{f}'(x)\right\}\right|\leq \frac{2}{m}\sqrt{\Vare\{\bar{f}(x)\}}\sqrt{\Vare\{\bar{f}'(x)\}}\\
	& \leq \frac{2}{m}\sqrt{\frac{1}{m}\sum_{k=1}^{m}v_k(x)}\sqrt{\frac{1}{m}\sum_{k=1}^{m}\Vare\{\widehat{f}_k'(x)\}}.
	\end{split}
	\]
	From the above two equations/inequalities, under conditions C4'(b)-(c), one has that
	\[
	\|\bar{v}\|_{\sup}\leq \frac{1}{m^2}\sum_{k=1}^{m}\|v_k\|_{\sup} \frac{1}{m}O_{\Px}(V_n),
	\]
	and that
	\[
	\begin{split}
	J_1(\bar{v})&=\int_{\mathcal{X}}\{\bar{v}_k'(x)\}^2\,d\Px(x)\bar{v}\leq \frac{4}{m^2}\int_{\mathcal{X}}\left\{\frac{1}{m}\sum_{k=1}^{m}v_k(x)\right\}^2\frac{\frac{1}{m}\sum_{k=1}^{m}\Vare\{\widehat{f}_k'(x)\}}{\frac{1}{m}\sum_{k=1}^{m}v_k(x)}\,d\Px(x)\\
	&\leq \frac{4}{m^2}\left\{\max_{1\le k\le m}\|v_k\|_{\sup}\right\}^2\int_{\mathcal{X}}\max_{1\le k\le m}\frac{\Vare\{\widehat{f}_k'(x)\}}{v_k(x)}\,d\Px(x)= \frac{1}{m^2} O_{\Px}(V_n^2H_n^2)
	\end{split}
	\]
	Therefore, under conditions C4'(a)-(b), we have that
	\[
	\bar{v}(x)\in \bar{\mathcal{F}}_n\text{ in probability measure $\Px$.}
	\]
	Using the facts that $Q=\Pb\bar{v}$ and $Q_1=\Pb_N\bar{v}$,  a direct application of~(\ref{eq:emp2}) gives that
	\[
	P\left(\frac{|Q_1-Q|}{Q_1+Q+\gamma_N}>C\epsilon\right)=P\left(\sup_{\bar{v}\in\bar{\mathcal{V}}_N}\frac{|\Pb_N \bar{v}-\Pb  \bar{v}|}{\Pb_N  \bar{v}+\Pb  \bar{v}+\gamma_N}>C\epsilon\right)\leq \exp(-N\epsilon^2\gamma_N/2)\to 0,
	\]
	where the last step follows from condition C4'(c). Furthermore, by conditions C4'(b)-(c), we have that $\frac{\gamma_N}{Q}=\sqrt{\frac{c_0(1+H_n)V_n}{mNQ^2}}=O_{\Px}(1)$. Hence we conclude that
	\be
	\label{eq:Q1}
	Q_1(\lambda|\Xv)=Q(\lambda|\Xv)+o_{\Px}\{Q(\lambda|\Xv)\}.
	\ee
	Combining equations (\ref{eq:Q2})--(\ref{eq:Q1}), we have that 
	\be
	\label{eq:Q1Q2}
	\frac{\frac{1}{Nm}\sum_{k=1}^m\tr(\bA_{kk}^2)}{\frac{\tr(\bar{\bA}_m^T\bar{\bA}_m)}{N}}=\frac{Q_1(\lambda|\Xv)}{Q_2(\lambda|\Xv)}=O_{\Px}(1).
	\ee
	By the definition of $\bar\bA_m$, it is straightforward to show that
	\[
	\begin{split}
	\frac{\{\frac{1}{N}\tr(\bar\bA_m)\}^2}{\frac{1}{Nm}\sum_{k=1}^m\tr(\bA_{kk}^2)}=\frac{1}{N}\frac{\{\frac{1}{m}\sum_{k=1}^m\tr(\bA_{kk})\}^2}{\frac{1}{m}\sum_{k=1}^m\tr(\bA_{kk}^2)}\leq \frac{1}{N}\frac{1}{m}\sum_{k=1}^m\frac{\{\tr(\bA_{kk})\}^2}{\tr(\bA_{kk}^2)}=\frac{1}{m}\sum_{k=1}^m\frac{\{N^{-1}\tr(\bA_{kk})\}^2}{N^{-1}\tr(\bA_{kk}^2)},
	\end{split}
	\]
	where the second last inequality follows from Cauchy-Schwartz inequality. Combining the above inequality and (\ref{eq:Q1Q2}), under condition C4'(a), we finally have that 
	\[
	\frac{\{\frac{1}{N}\tr(\bar\bA_m)\}^2}{\{\frac{1}{N}\tr(\bar\bA_m^T\bar\bA_m)\}}=\frac{\{\frac{1}{N}\tr(\bar\bA_m)\}^2}{\frac{1}{Nm}\sum_{k=1}^m\tr(\bA_{kk}^2)}\frac{\frac{1}{Nm}\sum_{k=1}^m\tr(\bA_{kk}^2)}{\{\frac{1}{N}\tr(\bar\bA_m^T\bar\bA_m)\}}=o_{\Px}(1),
	\]
	which completes the proof.
\end{proof}
\end{document}